%% file: main.tex
\documentclass[10pt,twocolumn,letterpaper]{article}

\usepackage[pagenumbers]{cvpr} 


\usepackage{graphicx}
\usepackage{caption}
\usepackage{wrapfig}
\usepackage{amsmath}
\usepackage{multirow}
\usepackage{fancybox}
\usepackage{adjustbox}
\usepackage{marginnote}
\usepackage{fancybox}
\usepackage{multirow}
\usepackage{multicol}
\usepackage{array}

\usepackage[resetlabels]{multibib}
\newcites{sup}{Reference}

\definecolor{cvprblue}{rgb}{0.21,0.49,0.74}
\usepackage[pagebackref,breaklinks,colorlinks,allcolors=cvprblue]{hyperref}

\newcommand{\best}[1]{\textcolor{red}{#1}}
\newcommand{\sbest}[1]{\textcolor{blue}{#1}}

\def\paperID{4655} 
\def\confName{CVPR}
\def\confYear{2025}

\title{Robust Message Embedding via Attention Flow-Based Steganography}

\author{Huayuan Ye$^1$, Shenzhuo Zhang$^1$, Shiqi Jiang$^1$, Jing Liao$^2$, Shuhang Gu$^3$, Dejun Zheng$^4$, \\
Changbo Wang$^1$, Chenhui Li$^{1*}$\\
$^1$ East China Normal University~~~~~~~~~$^2$ City University of Hong Kong\\
$^3$ University of Electronic Science and Technology of China, Chengdu, China\\
$^4$ Zhijiang College of Zhejiang Univeristy of Technology, Zhejiang, China\\
{\tt\small huayuan221@gmail.com; \{10195102459, 52265901032\}@stu.ecnu.edu.cn; jingliao@cityu.edu.hk;}\\
{\tt\small shuhanggu@gmail.com; zhengdejun@zzjc.edu.cn; \{cbwang, chli\}@cs.ecnu.edu.cn}
}

\begin{document}
\maketitle
\input{sec/abstract.tex}    
\input{sec/introduction.tex}
\input{sec/relatedwork.tex}

\input{sec/method.tex}
\input{sec/experiment.tex}
\input{sec/conclusion.tex}
{
    \small
    \bibliographystyle{ieeenat_fullname}
    \bibliography{main}
}

\input{sec/appendix.tex}
\clearpage
{
    \small
    \renewcommand\refnamesup{References}
    \bibliographystylesup{ieeenat_fullname}
    \bibliographysup{sup}
}

\end{document}

%% file: sec/abstract.tex
\begin{abstract}
    Image steganography can hide information in a host image and obtain a stego image that is perceptually indistinguishable from the original one. This technique has tremendous potential in scenarios like copyright protection, information retrospection, etc. Some previous studies have proposed to enhance the robustness of the methods against image disturbances to increase their applicability. However, they generally cannot achieve a satisfying balance between the steganography quality and robustness. Instead of image-in-image steganography, we focus on the issue of message-in-image embedding that is robust to various real-world image distortions. This task aims to embed information into a natural image and the decoding result is required to be completely accurate, which increases the difficulty of data concealing and revealing. Inspired by the recent developments in transformer-based vision models, we discover that the tokenized representation of image is naturally suitable for steganography task. In this paper, we propose a novel message embedding framework, called \textbf{R}obust \textbf{M}essage \textbf{Steg}anography~(RMSteg), which is competent to hide message via QR Code in a host image based on an normalizing flow-based model. The stego image derived by our method has imperceptible changes and the encoded message can be accurately restored even if the image is printed out and photoed. To our best knowledge, this is the first work that integrates the advantages of transformer models into normalizing flow. Our experiment result shows that RMSteg has great potential in robust and high-quality message embedding.
\end{abstract}

%% file: sec/introduction.tex
\section{Introduction}

\begin{figure}[htb]
  \centering
  \includegraphics[width=1.0\linewidth]{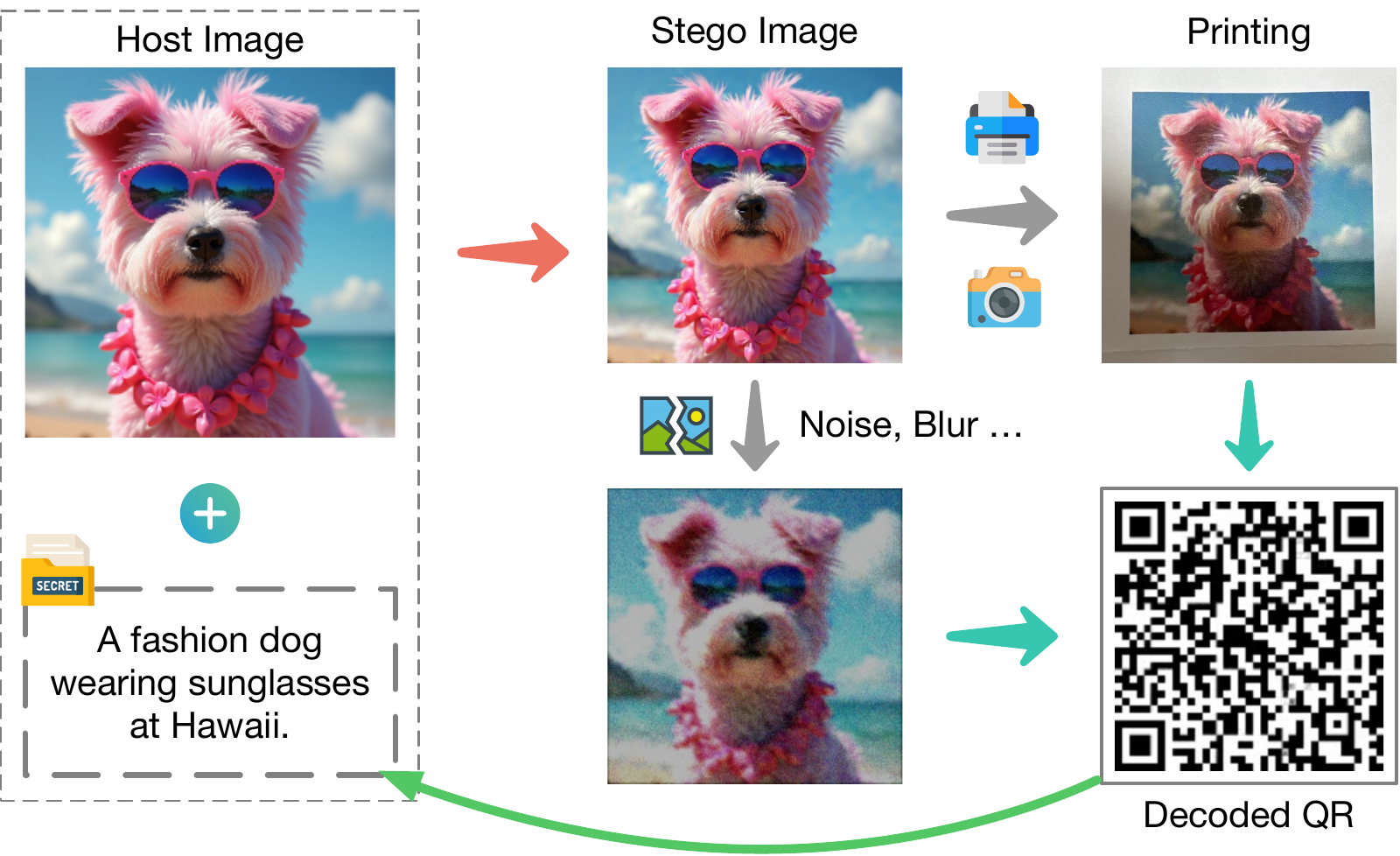}
  \caption{Compared with previous methods that can only embed limited bit-level information, RMSteg can achieve a much higher embedding capacity and meanwhile has better steganography quality. Also, it can survive various real-world distortions.
  }
  \label{fig: teaser}\vspace{-10pt}
\end{figure}

Steganography, the art of hiding secret information in a carrier, has long been a hot research topic. This technique is competent to embed information like images and text into target containers, thus achieving copyright protection~\cite{fu2022chartstamp,tancik2020stegastamp}, information retrospection~\cite{ye2023invvis, zhang2020viscode}, etc. Steganography aims to prevent people from discovering the existence of secret data instead of understanding the meaning of data, differentiating it from cryptography. Specifically, image steganography uses image as the carrier to hide secret information.

Traditional image steganography methods mainly modify the image in spatial domain~\cite{imaizumi2014multibit, kawaguchi1999principles, mielikainen2006lsb, nguyen2006multi, niimi2002high} or transform domain~\cite{almohammad2008high, swanson1997multiresolution, zhu1999multiresolution}. This kind of method is easy to be detected by steganalysis techniques~\cite{fridrich2001detecting, yu2004reliable}, which makes it lack security. Recently, with the developments of deep learning, some deep steganography methods have been proposed. Most of them are based on autoencoder~\cite{baluja2017hiding, wengrowski2019light, zhang2020viscode, zhu2018hidden} and normalizing flow~\cite{cheng2021iicnet, guan2022deepmih, jing2021hinet, lu2021large, xu2022robust, ye2023invvis}.

Images can undergo various digital or real-world disturbances during dissemination. To enable the stego images to survive these distortions, some robust steganography methods have been proposed. They consider various distortion situations like light field messaging~(LFM)~\cite{wengrowski2019light}, JPEG compression~\cite{xu2022robust}, etc. In the field of robust steganography, robust message embedding is very promising in many application scenarios like hyperlink hiding~\cite{tancik2020stegastamp}, metadata embedding~\cite{hota2019embedding, zhang2020viscode}, etc. However, this task requires the decoding result being completely accurate with no error, which poses a challenge to the balance between stego image quality and decoding accuracy, especially when facing real-world distortions. Although some studies~\cite{fu2022chartstamp, tancik2020stegastamp} have proposed to hide messages in host images and try to make them survive printing and photography, which is among the most demanding situations that require high steganography robustness, they cannot achieve enough steganography quality and capacity at the same time.

As the most widely utilized method, normalizing flow-based model~\cite{dinh2014nice, dinh2016realnvp, kingma2018glow} has achieved impressive performance in various steganography tasks. Existing methods~\cite{cheng2021iicnet, guan2022deepmih, jing2021hinet, lu2021large, xu2022robust, ye2023invvis} generally incorporate normalizing flow by utilizing a convolutional neural network~(CNN) based backbone. However, according to our experiments, this kind of model design can lead to obvious artifacts in stego images when handling robust steganography tasks due to the lack of inner-channel feature fusion. Inspired by the transformer-based vision models, we discover that the tokenized representation of image is naturally suitable for robust steganography that requires highly abstract feature learning. As a result, we aim to take advantages of it to address the robust message-in-image steganography problem.

In this paper, we propose a new framework for message embedding, called \textbf{R}obust \textbf{M}essage \textbf{Steg}anography~(RMSteg), a simple demo is demonstrated in \autoref{fig: teaser}. We use QR Code as the message carrier and encode it into the host image. Unlike previous methods that directly encode the secret image, we propose an invertible QR Code transition as a preprocessing step, which transforms the QR Code based on the features of the host image, lowering down the artifacts in stego images and meanwhile maintaining a high decoding accuracy. We outline an steganography model called AttnFlow, which integrates tokenized image representation into normalizing flow. We propose an attention affine coupling block~(AACB) that leverages the attention mechanism~\cite{vaswani2017attention, dosovitskiy2020image} instead of traditional CNN for invertible steganography function learning, thus significantly improving the stego image quality. Compared with previous methods, our method can conquer the aforementioned difficulties and achieve robust, high-quality and high-capacity message-in-image steganography. The main contributions of this paper include three aspects:

\begin{itemize}[leftmargin=0.15in]
    \item We use QR Code as the message carrier and propose a transition scheme to transform the QR Code before steganography. This process can improve the stego image quality while maintaining decoding accuracy.
    \item We propose an invertible token fusion module that can effectively improve the steganography quality by simply including a small learnable matrix.
    \item We propose a normalizing flow-based steganography network that integrates the tokenized image representation. Our network can generate stego images with significantly higher quality and can survive extreme distortions. We use the case of printing and photography to validate our method's effectiveness.
\end{itemize}

%% file: sec/relatedwork.tex
\section{Related Work}

\subsection{Image Steganography}
\textbf{Traditional Steganography} Image steganography hides information in an image by performing imperceptible changes on a host image. Traditional methods modify the image in spatial or transform domain~\cite{baluja2017hiding}. Spatial-domain steganography generally leverages least-significant-bit~(LSB) replacement~\cite{mielikainen2006lsb}, bit plane complexity segmentation~(BPCS)~\cite{kawaguchi1999principles, nguyen2006multi} and palette reordering~\cite{imaizumi2014multibit, niimi2002high} to conceal information. However, this kind of scheme may raise statistical anomalies that can be detected by steganalysis techniques~\cite{fridrich2001detecting, yu2004reliable}. Some methods utilize high-dimensional features~\cite{pevny2010using} and distortion constraints~\cite{li2014new} to improve steganography security and quality. Transform-based steganography can hide data in a transformed domain using discrete cosine transform~(DCT)~\cite{almohammad2008high} and discrete wavelet transform~(DWT)~\cite{swanson1997multiresolution, zhu1999multiresolution}. Due to the limited ability of feature representation and transformation, traditional methods generally cannot achieve a satisfying quality.

\textbf{Deep Steganography} Recently, various deep learning-based image steganography schemes have been proposed and have achieved impressive performance. HiDDeN~\cite{zhu2018hidden} adopted the autoencoder~(AE) to embed binary messages. Baluja~\cite{baluja2017hiding} first utilized an end-to-end network to hide a color image in another. Some studies~\cite{fu2022chartstamp, qin2020coverless, shi2017ssgan, tang2019cnn, tang2017automatic, zhang2019steganogan} incorporated generative adversarial network~(GAN)~\cite{goodfellow2014generative} to reduce the image artifacts and  defend steganalysis. More recently, the invertible neural network~(INN)~\cite{dinh2014nice, dinh2016realnvp, kingma2018glow} has been widely used for steganography. These methods successfully hide single~\cite{cheng2021iicnet, jing2021hinet, lu2021large} or multiple~\cite{guan2022deepmih, wu2021embedding, ye2023invvis} images in a carrier image. There are also some studies focusing on coverless steganography~\cite{li2022coverless, liu2020coverless, mohamed2021coverless, yu2024cross} that directly transforms the secret information into a cover image. These methods mainly focus on improving the embedding capacity instead of robustness, as a result, they generally cannot survive image distortions.

\textbf{Robust Steganography} Robust steganography allows information decoding even if the images are interfered with by digital transmission or real-world distortions, which is meaningful for scenarios like copyright protection, secret communication, etc. VisCode~\cite{zhang2020viscode} hides QR Codes in host images and can survive image brightness changes and slight tampering. LFM~\cite{wengrowski2019light} is robust to light field messaging. RIIS~\cite{xu2022robust} considers JPEG compression and various kinds of noise separately based on a conditional network. StegaStamp~\cite{tancik2020stegastamp} and ChartStamp~\cite{fu2022chartstamp} take printing and photography into account but can only embed very little information at a cost of visual quality loss. As far as we know, existing methods cannot achieve both high-quality and high-capacity message steganography that is robust to extreme image distortions. In this paper, we aim to take the advantages of transformer-based model to address this problem.

\subsection{Normalizing Flow-Based Models}
Normalizing flow model was first proposed as a generative model by Dinh~et~al.~\cite{dinh2014nice}. With further improvement by RealNVP~\cite{dinh2016realnvp} and GLOW~\cite{kingma2018glow}, it is also known as the invertible neural network (INN). INN can learn an invertible function using a set of affine coupling layers with shared parameters to map the original data distribution to a simple distribution (e.g., Gaussian distribution). Chen~et~al.~\cite{chen2019residual} proposed an unbiased estimation for normalizing flow model. i-RevNet~\cite{jacobsen2018revnet} utilizes an explicit inversion to improve the invertible architecture.

Recently, normalizing flow has been applied to various downstream tasks in computer vision, such as image~\cite{xiao2020invertible} and video~\cite{zhu2019residual} super-resolution, image-to-image translation~\cite{van2019reversible}, etc. Especially, in the field of steganography, normalizing flow-based methods~\cite{cheng2021iicnet, lu2021large,ye2023invvis} have shown promising performance. HiNet~\cite{jing2021hinet} introduces the discrete wavelet transform~(DWT) to guide channel squeezing and improve the steganography quality. DeepMIH~\cite{guan2022deepmih} hides single or multiple images with a saliency detection module. Mou~et~al.~\cite{mou2023large} incorporated a key-controllable network design to implement secure video steganography. Xu~et~al.~\cite{xu2022robust} simulated distortions during model training to improve the robustness and security of their method. Although previous studies have leveraged various methods to improve the network architecture for better performance, they cannot attend to both image quality and steganography robustness simultaneously.

%% file: sec/method.tex
\section{Method}
\begin{figure*}[t]
    \centering
    \includegraphics[width=1\linewidth]{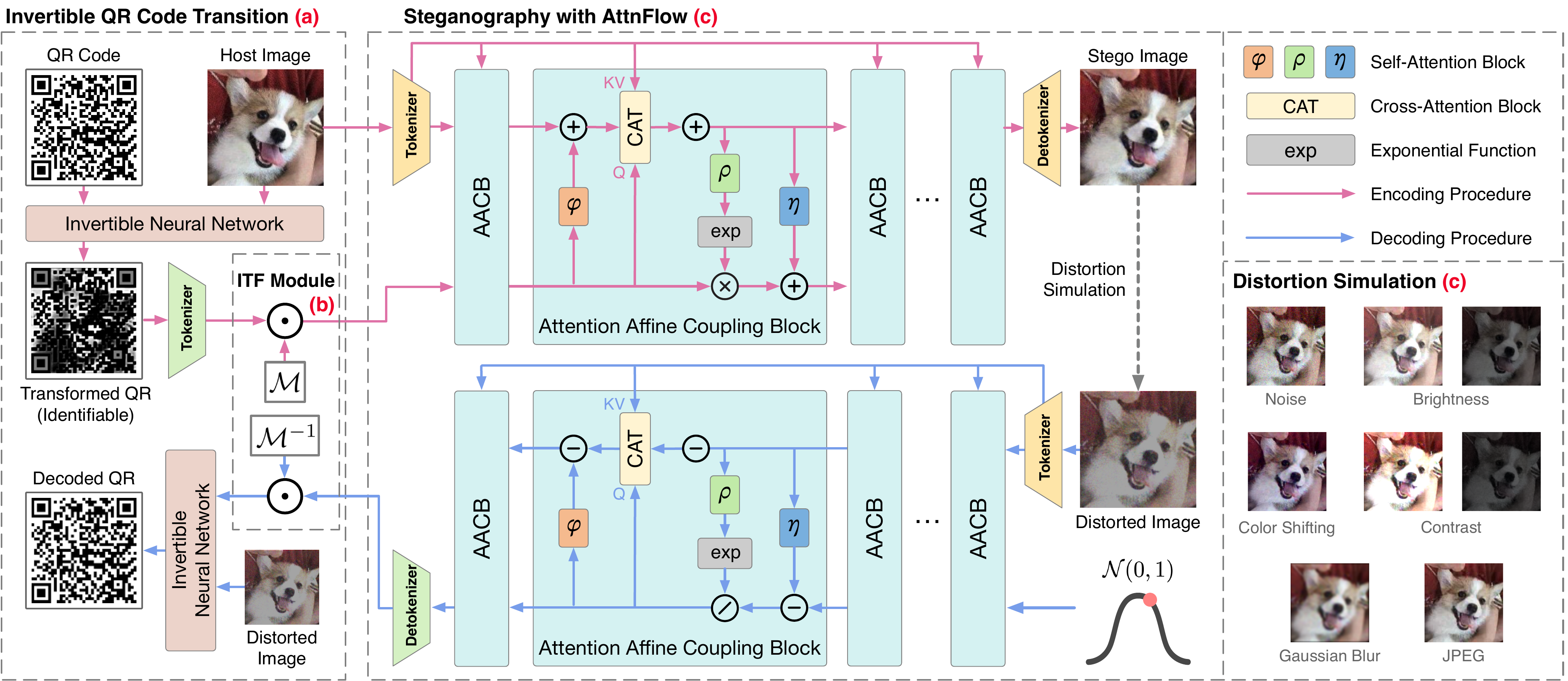}\vspace{-6pt}
    \caption{
        The pipeline of RMSteg. We first transform the QR Code encoded with the secret message to make it easier to hide through an invertible neural network~(a). After that, we perform invertible token fusion~(ITF) (b) on the tokenized QR Code. We then use a normalizing flow-based model with attention affine coupling blocks~(AACBs) to implement data concealing and revealing~(c). During training, we employ a distortion simulation module~(d) to simulate real-world image disturbances.
   }
  \label{fig: model}\vspace{-12pt}
\end{figure*}

\subsection{Overview}

Given a secret message $T_s$, we first encode it into a QR Code image $I_q$. The concealing procedure aims to embed $I_q$ into a host image $I_h$ and derive a stego image $I_s$ that is perceptually similar to $I_h$. Then, $I_s$ can suffer from various real-world image distortions, resulting in a distorted image $I_s'$. After that, the revealing procedure aims to restore a QR Code $\hat{I}_q$ from $I_s'$ that can be successfully recognized to obtain the original message.

To achieve the aforementioned targets, we first leverage a QR Code transition scheme~(Sec.~\ref{sec: qr_trans}) to transform the original QR Code according to the host image, reducing the artifacts it causes in the subsequent steganography process. Then, we use an invertible token fusion~(ITF) module~(Sec.~\ref{sec: itf}) to improve the stego image quality. After that, we propose an AttnFlow model~(Sec.~\ref{sec: attnflow}) to perform message embedding. To make our method robust to real-world distortions, we incorporate a distortion simulation module during the training stage, which will be described in detail in Sec.~\ref{sec: optim}. Fig.~\ref{fig: model} demonstrates an overview of the pipeline of our RMSteg.

\subsection{Preliminary: Normalizing Flow}
Normalizing Flow~\cite{dinh2014nice, dinh2016realnvp, kingma2018glow}, also called the invertible neural network~(INN), is proposed to model a bijective projection from a complex distribution~(e.g., images) to a tractable distribution~(e.g. Gaussian distribution and Dirac distribution). This kind of model generally comprises several invertible affine coupling blocks~(ACBs). The most basic ACB architecture is proposed by NICE~\cite{dinh2014nice}, in which the input $u^{i}$ of the $i$\textsuperscript{th} ACB is split into two parts, $u^{i}_1$ and $u^{i}_2$, whose corresponding outputs are $u^{i + 1}_1$ and $u^{i + 1}_2$, respectively. For the forward process, the following transformation is performed:
\begin{equation}\label{eq: inn_forward}
    u^{i + 1}_1 = u^{i}_1 + \sigma(u^{i}_2),~~u^{i + 1}_2 = u^{i}_2 + \delta(u^{i + 1}_1),
\end{equation}
where $\sigma(\cdot)$ and $\delta(\cdot)$ are arbitrary functions. Obviously, the backward process can be formulated as:
\begin{equation}\label{eq: inn_backward}
    u^{i}_2 = u^{i + 1}_2 - \delta(u^{i + 1}_1),~~u^{i}_1 = u^{i + 1}_1 - \sigma(u^{i}_2).
\end{equation}
In the normalizing flow architecture, $\delta(\cdot)$ and $\sigma(\cdot)$ in \autoref{eq: inn_forward} and \autoref{eq: inn_backward} can be implemented by neural network modules with shared parameters and inverse calculation manner. By stacking multiple ACBs, the network can learn an invertible transformation between two distributions. Since this scheme is inherently suitable for steganography, many studies have utilized it for data hiding and proposed various improvements. In this paper, we further extend the ability of normalizing flow and propose a new network architecture for our robust message embedding task.

\subsection{Invertible QR Code Transition}
\label{sec: qr_trans}

For the message embedding task in this paper, the hidden QR code needs to be restored with enough accuracy to be identified by common devices like cell phones, webcams, etc. To balance the trade-off between the stego image quality and decoding accuracy, VisCode~\cite{zhang2020viscode} obtains a visual saliency map to guide the QR Code embedding while ChartStamp~\cite{fu2022chartstamp} utilizes the semantic segmentation result as the training loss guidance. Although this kind of rule-based strategy can improve the visual quality of the stego image, it does not consider the inherent relationship between QR Codes and the host image.

In our method, we adopt a more direct approach, which is modifying the QR Code image according to the host image (shown in \autoref{fig: model}~(a)). We call it invertible QR Code transition~(IQRT). The key idea of IQRT is that, the QR Code used for steganography is not necessarily to be black-and-white to keep its information. Thus, a learnable transformation can be applied to the QR Code for a better steganography quality as long as the transformed code is still identifiable. Formally, given a host image $I_h$ and a QR Code $I_q$ with the same size, we use an off-the-shelf INN architecture\footnote{We only use 2 invertible blocks instead of 16 in the original paper~\cite{lu2021large}.} proposed by ISN~\cite{lu2021large} to learn an invertible function $f(\cdot)$ that derives the transformed QR Code $I_q^*$ by $I_q^* = f(I_q, I_h)$. In the reverse process, the restored QR Code $\hat{I_q}$ can be obtained by $\hat{I_q} = f^{-1}(I_q^*, I_s')$, where $f^{-1}(\cdot)$ is the inverse function of $f(\cdot)$ defined by normalizing flow and $I_s'$ is the distorted stego image. Here $I_s'$ is used instead of $I_h$ since the latter is unknown in the decoding procedure.

\begin{figure}[]
    \centering
    \includegraphics[width=1.0\linewidth]{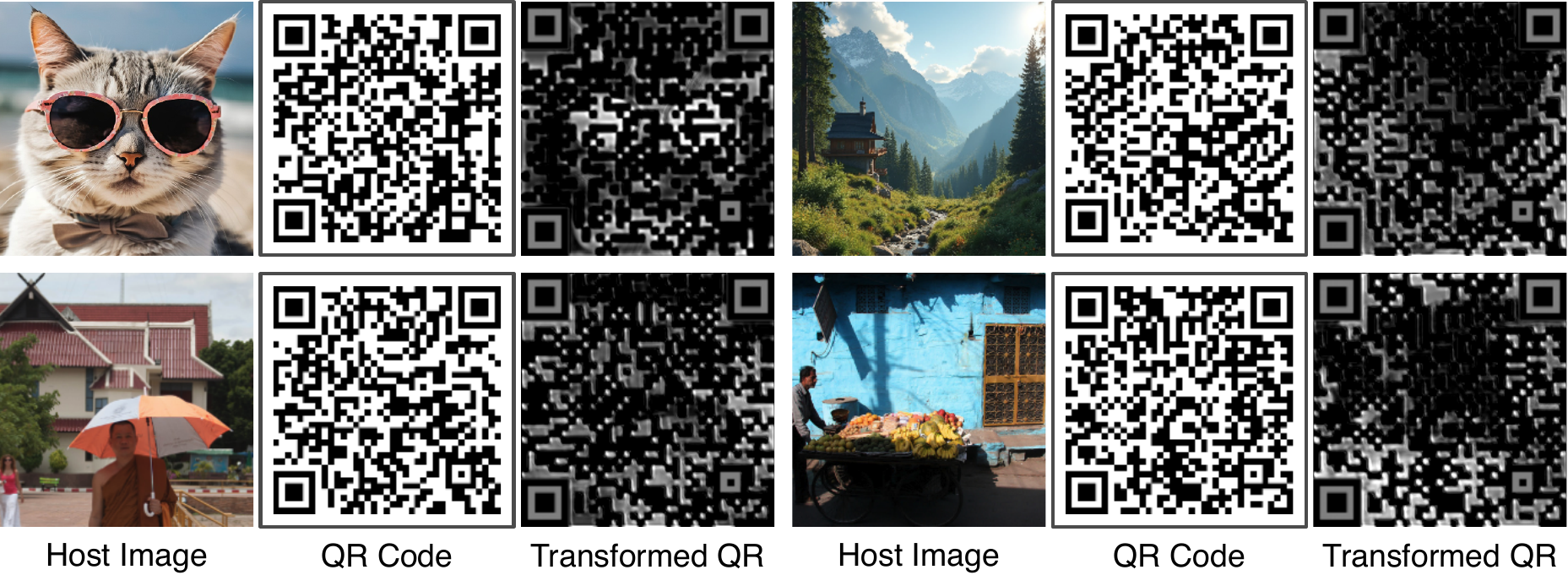}\vspace{-8pt}
    \caption{Some QR Code transition results, the transformed QR Codes are still identifiable.}
    \label{fig: qr_trans}
    \vspace{-13pt}
\end{figure}

During network training, the transition network is jointly trained with the subsequent steganography network. We employ the same constraint as ArtCoder~\cite{su2021artcoder} to the transformed QR Code to ensure that it is still identifiable. Specifically, a Gaussian convolution kernel is applied to each code module to simulate the QR Code scanning procedure. For more details\footnote{A detailed explanation is also provided in the appendix.}, we suggest referring to the original paper~\cite{su2021artcoder}. We do not use extra constraint to the transition network so that it can learn the best transition strategy according to the overall optimization targets. \autoref{fig: qr_trans} shows some transition results, it can be observed that the transformed QR Codes have obviously lower brightness. However, with the aforementioned constraint, the transformed QR Codes are still identifiable, guaranteeing almost no information loss.

\subsection{Invertible Token Fusion}
\label{sec: itf}

With the transformed QR Code, we first use a ViT~\cite{dosovitskiy2020image} to obtain a tokenized representation $T_q \in \mathbb{R}^{N \times D}$, in which $N$ is the number of tokens and $D$ represents the token dimensionality. Inspired by the invertible $1 \times 1$ convolution proposed by GLOW~\cite{kingma2018glow}, before feeding $T_q$ to the subsequent steganography network, we put forth an invertible token fusion~(ITF) module (as shown in \autoref{fig: model}~(b)) to transform the QR Code tokens for a better steganography quality.

Formally, we use a learnable matrix $\mathcal{M} \in \mathbb{R}^{N \times N}$, which is initialized as an orthogonal matrix using Cholesky decomposition~\cite{krishnamoorthy2013matrix}, as a transform matrix for $T_q$. In the steganography process, $T_q$ is transformed by performing a matrix multiplication: $T_q' = \mathcal{M} \cdot T_q$. Obviously, in the decoding procedure, the restored tokens $\hat{T_q}$ can be obtained by: $\hat{T}_q = \mathcal{M}^{-1} \cdot \hat{T}_q'$, where $\mathcal{M}^{-1}$ is the inverse matrix.

Different from GLOW~\cite{kingma2018glow} that utilizes the invertible convolution to learn the channel-wise fusion strategy, our ITF module learns a patch-wise transformation that enables inner-channel feature interaction. Our experiments also prove that ITF can efficiently and effectively improve the steganography quality by simply introducing the aforementioned learnable matrix.

\subsection{Steganography with AttnFlow}
\label{sec: attnflow}

Previous steganography studies based on normalizing flow generally adopt a convolutional neural network~(CNN) based backbone, mostly DenseNet~\cite{wang2018esrgan}, to construct the affine coupling blocks~(ACBs). This kind of design only considers the channel-wise feature fusion and can lead to perceptible artifacts in stego images, especially in the robust steganography task. Motivated by the impressive performance achieved by the transformer-based~\cite{dosovitskiy2020image, vaswani2017attention} vision models recently, we propose a model called AttnFlow that introduces attention mechanism to normalizing flow to implement robust steganography.


As shown in \autoref{fig: model}~(c), similar to ordinary normalizing flow, AttnFlow contains several attention affine coupling blocks~(AACBs) for invertible function learning. Assume that the input of the $i$\textsuperscript{th} AACB is split into $T_h^{(i - 1)}$ and $T_q^{(i - 1)}$, corresponding to the host image tokens and QR Code tokens, respectively. Specifically, $T_h^{(0)}$ is the tokenized host image obtained with a basic ViT~\cite{dosovitskiy2020image} and $T_q^{(0)}$ represents the QR Code image tokens output by the ITF module. For the $i$\textsuperscript{th} AACB, we perform the following affine transformation:
\begin{equation}\label{eq: aacb_forward}
    \begin{split}
        T_h^{(i)} &= T_h^{(i - 1)} + \phi(T_q^{(i - 1)}) + \mathcal{C}(T_q^{(i - 1)}, T_h^{(0)}) \times \alpha_i, \\
        T_q^{(i)} &= \eta(T_h^{(i)}) + T_q^{(i - 1)} \odot exp(\rho(T_h^{(i)})),
    \end{split}
\end{equation}
in which $\phi(\cdot)$, $\eta(\cdot)$, $\rho(\cdot)$ are self-attention blocks~\cite{vaswani2017attention} followed by a feedforward multilayer perceptron~(MLP), $\mathcal{C}(q, kv)$ represents the cross-attention block~\cite{vaswani2017attention}, $exp(\cdot)$ is the exponential function, $\odot$ indicates the Hadamard product and $\alpha_i$ is a dependent trainable coefficient for each AACB. We calculate the attention value with:
\begin{equation}
    Attn(Q, K, V) = M \cdot V,~~M = Softmax(\frac{QK^T}{\sqrt{d}}),
\end{equation}
where $Q$, $K$, $V$ are derived from the learned projections and $d$ is the dimension of the projected tokens. As described in \autoref{eq: aacb_forward}, in addition to the self-attention value, we also calculate the cross-attention value of the initial host image tokens $T_h^{(0)}$ upon the QR code tokens $T_q^{(i - 1)}$ for each AACB. Then, we add these values to $T_h^{(i - 1)}$ to help AACBs gradually integrate the information from the QR code into the image. For the QR Code tokens, we choose to adopt a generally incorporated~\cite{cheng2021iicnet, jing2021hinet, lu2021large, ye2023invvis} affine transformation and replace the original convolutional blocks with $\eta(\cdot)$ and $\rho(\cdot)$. After $n$ AACBs, $T_h^{(n)}$ further goes through a detokenizer\footnote{The detailed architecture of the tokenizers and detokenizers will be described in the appendix.}, resulting in the final stego image. Although some methods further map $T_h^{(n)}$ and $T_q^{(n)}$ as a conditional distribution for better performance, here we choose to adopt the same assumption as HiNet~\cite{jing2021hinet}, which is simply positing that $T_q^{(n)}$ obeys a Gaussian distribution.

In the revealing process, we aim to restore the original QR Code from a distorted stego image $I_s'$, we first tokenize it and derive $\hat{T}_h^{(n)}$. Then, we obtain $\hat{T}_q^{(n)}$ by sampling from a standard Gaussian distribution. After that, we perform the inverse AACB transformation by going through them with inverse calculation manner:
\begin{equation}\label{eq: aacb_backrward}
    \begin{split}
        \hat{T}_q^{(i - 1)} &= (\hat{T}_q^{(i)} - \eta(\hat{T}_h^{(i)})) \odot exp(-\rho(\hat{T}_h^{(i)})), \\
        \hat{T}_h^{(i - 1)} &= \hat{T}_h^{(i)} - \phi(\hat{T}_q^{(i - 1)}) - \mathcal{C}(\hat{T}_q^{(i - 1)}, \hat{T}_h^{(0)}) \times \alpha_i,
    \end{split}
\end{equation}
in which $\hat{T}_h^{(0)}$ is obtained by tokenizing $I_s'$ since $I_h$ is unknown in the decoding process. Then, $\hat{T}_q^{(0)}$ is detokenized and fed into the reversed QR Code transition (introduced in Sec.~\ref{sec: qr_trans}) with $I_s'$ to get the final decoded QR Code image.

\subsection{Optimization Target and Training Strategy}
\label{sec: optim}

\textbf{Distortion Simulation Module} We use a module to simulate the distortions that the stego images may undergo during printing and photography. In this paper, we choose to use the same simulation module proposed by StegaStamp~\cite{tancik2020stegastamp} that considers color shifting, blurring, noising, etc. We mainly modify the standard deviation of Gaussian noise from 0.02 to 0.07 and increase the JPEG compression quality from 25 to 60 for our task. During training, we perform random distortion combinations on stego images to simulate real-world image disturbances.

\textbf{Loss Function} The aforementioned three networks (IQRT, ITF and AttnFlow) are trained jointly. We use the following loss functions to guide the training process:
\begin{equation}
    \mathcal{L}_{steg}^{L1} = \left\| I_h - I_s \right\|_1,
\end{equation}
\vspace{-12pt}
\begin{equation}
    \mathcal{L}_{steg}^{ssim} = ssim(I_h, I_s),
\end{equation}
\vspace{-10pt}
\begin{equation}
    \mathcal{L}_{steg}^{lpips} = lpips(I_h, I_s),
\end{equation}
\vspace{-10pt}
\begin{equation}
    \mathcal{L}_{qr} = \| I_q - \hat{I}_q \|_1,
\end{equation}
in which $ssim(\cdot)$ represents the structural similarity index~\cite{wang2004image} and $lpips(\cdot)$ indicates the perception loss~\cite{zhang2018unreasonable}. Besides, as introduced in Sec.~\ref{sec: qr_trans}, an additional QR Code transition loss $\mathcal{L}_{t}$ is incorporated. The overall loss function is the weighted sum of the above functions:
\begin{equation}\label{eq: loss}
    \mathcal{L}_{total} = \alpha \mathcal{L}_{steg}^{L1} + \beta \mathcal{L}_{steg}^{ssim} + \gamma \mathcal{L}_{steg}^{lpips} + \delta \mathcal{L}_{qr} + \epsilon \mathcal{L}_{t},
\end{equation}
where $\alpha$, $\beta$, $\gamma$, $\delta$, $\epsilon$ are weight coefficients.


%% file: sec/experiment.tex
\begin{figure}[t]
    \centering
    \includegraphics[width=1.0\linewidth]{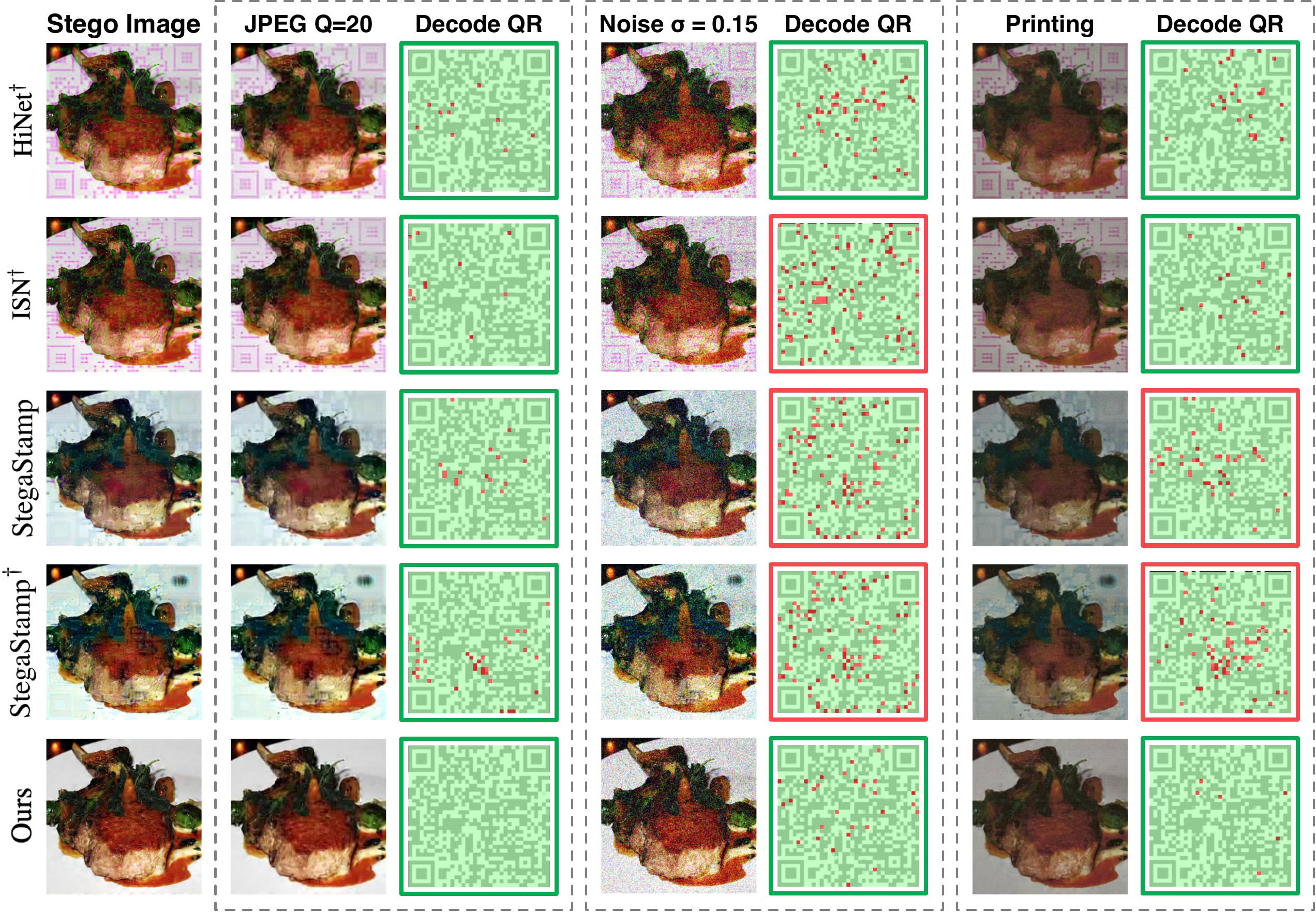}
    \vspace{-20pt}
    \caption{
        Stego images and decoded QR Codes under different distortions. QR Codes with green borders can be recognized while those with red borders cannot. Zoom in for better observation.
   }
  \label{fig: evl_quality} 
  \vspace{-12pt}
\end{figure}

\begin{table*}[htb]
    \caption{
        Steganography quality under different situations. Here $\sigma$ represents the standard deviation of Gaussian noise~(given the image pixel values range in $[0, 1]$). The best and second-best results are marked in \textcolor{red}{red} and \textcolor{blue}{blue} colors, respectively.
    }\vspace{-10pt}
    \newcolumntype{M}[1]{>{\centering\arraybackslash}m{#1}}
    \renewcommand\arraystretch{0.7}
    \centering
    \small
    \scalebox{1.01}{
    \begin{tabular}{M{1.2cm}M{0.65cm}M{0.65cm}M{0.65cm}M{0.6cm}M{0.6cm}M{0.6cm}M{0.6cm}M{0.6cm}M{0.6cm}M{0.6cm}M{0.6cm}M{0.6cm}M{0.6cm}M{0.6cm}M{0.6cm}} 
    \bottomrule
    \multirow{2}{=}{\centering{Method}} & \multicolumn{3}{c}{\scriptsize{Stego Image}} & \multicolumn{2}{c}{\scriptsize{$\sigma = 0.1$}} & \multicolumn{2}{c}{\scriptsize{$\sigma = 0.15$}} & \multicolumn{2}{c}{\scriptsize{JPEG Q = 20}} & \multicolumn{2}{c}{\scriptsize{JPEG Q = 40}} & \multicolumn{2}{c}{\scriptsize{Mixed}} & \multicolumn{2}{c}{\scriptsize{Printing}}\\ [-0.2pt]
    & \scriptsize{PSNR$\uparrow$} & \scriptsize{SSIM$\uparrow$} & \scriptsize{LPIPS$\downarrow$} & \scriptsize{TRA$\uparrow$} & \scriptsize{EMR$\downarrow$} & \scriptsize{TRA$\uparrow$} & \scriptsize{EMR$\downarrow$} & \scriptsize{TRA$\uparrow$} & \scriptsize{EMR$\downarrow$} & \scriptsize{TRA$\uparrow$} & \scriptsize{EMR$\downarrow$} & \scriptsize{TRA$\uparrow$} & \scriptsize{EMR$\downarrow$} & \scriptsize{TRA$\uparrow$} & \scriptsize{EMR$\downarrow$} \\ [-0.2pt]
    \hline

    \scriptsize{~~ISN\textsuperscript{$\dagger$}} & \scriptsize{\sbest{32.175}} & \scriptsize{\sbest{0.8765}} & \scriptsize{0.3266} & \scriptsize{0.728} & \scriptsize{1.563} & \scriptsize{0.178} & \scriptsize{5.020} & \scriptsize{\sbest{0.991}} & \scriptsize{0.721} & \scriptsize{\sbest{0.999}} & \scriptsize{0.184} & \scriptsize{\sbest{0.713}} & \scriptsize{3.131} & \scriptsize{0.960} & \scriptsize{\sbest{1.125}} \\ [-0.2pt]

    \scriptsize{~~HiNet\textsuperscript{$\dagger$}} & \scriptsize{31.629} & \scriptsize{0.8662} & \scriptsize{0.3423} & \scriptsize{\best{0.827}} & \scriptsize{\best{1.077}} & \scriptsize{\sbest{0.162}} & \scriptsize{\sbest{3.724}} & \scriptsize{0.986} & \scriptsize{\sbest{0.573}} & \scriptsize{0.997} & \scriptsize{\sbest{0.099}} & \scriptsize{0.677} & \scriptsize{3.426} & \scriptsize{\sbest{0.970}} & \scriptsize{1.619} \\ [-0.2pt]

    \scriptsize{StegaStamp} & \scriptsize{21.215} & \scriptsize{0.7027} & \scriptsize{\sbest{0.3055}} & \scriptsize{0.051} & \scriptsize{6.152} & \scriptsize{0.000} & \scriptsize{10.57} & \scriptsize{0.951} & \scriptsize{1.259} & \scriptsize{0.977} & \scriptsize{0.798} & \scriptsize{0.557} & \scriptsize{3.843} & \scriptsize{0.750} & \scriptsize{3.214} \\ [-0.2pt]

    \scriptsize{StegaStamp\textsuperscript{$\dagger$}} & \scriptsize{21.173} & \scriptsize{0.6903} & \scriptsize{0.3418} & \scriptsize{0.481} & \scriptsize{3.298} & \scriptsize{0.015} & \scriptsize{6.500} & \scriptsize{0.953} & \scriptsize{1.104} & \scriptsize{0.969} & \scriptsize{0.833} & \scriptsize{0.693} & \scriptsize{\sbest{2.975}} & \scriptsize{0.900} & \scriptsize{1.917} \\ [-0.2pt]
    
    \scriptsize{~~Ours~~~} & \scriptsize{\best{32.883}} & \scriptsize{\best{0.9109}} & \scriptsize{\best{0.0707}} & \scriptsize{\sbest{0.794}} & \scriptsize{\sbest{1.235}} & \scriptsize{\best{0.216}} & \scriptsize{\best{3.306}} & \scriptsize{\best{0.995}} & \scriptsize{\best{0.117}} & \scriptsize{\best{1.000}} & \scriptsize{\best{0.038}} & \scriptsize{\best{0.859}} & \scriptsize{\best{0.861}} & \scriptsize{\best{1.000}} & \scriptsize{\best{0.606}} \\ [-0.2pt]

    \toprule
    \end{tabular}
    }
    \label{tab: evl_steg_quality}
    \vspace{-18pt}
\end{table*}

\section{Experiment}
\label{sec: eval}

\subsection{Experimental Settings}

\textbf{Datasets} Our training and testing datasets of host images are the \textsl{train2017}~(118K) and \textsl{test2017}~(41K) datasets of COCO~\cite{lin2014microsoft}, respectively. For QR Code images, we manually construct the training~(50K) and testing~(41K) datasets with random encoded messages. We generate the QR Code images by adopting the scheme of QR Code version 5~\cite{qrspec} with highest error correction~(ECC) level of `H'. We incorporate this code version for most of our experiments except the evaluation in Sec.~\ref{evl: capacity}. The image used for training and testing is 224 $\times$ 224 and the patch size of ViT~\cite{dosovitskiy2020image} is 16.


\textbf{Metrics} Our experiments focus on two aspects: stego image quality and decoding accuracy. For stego image quality, we use the peak signal-to-noise ratio~(PSNR), SSIM~\cite{wang2004image} and LPIPS~\cite{zhang2018unreasonable} to measure the difference between host images and stego images. For decoding accuracy, we adopt the text recovery accuracy~(TRA)~\cite{ye2023invvis,zhang2020viscode}, which is the ratio of the successfully decoded QR Codes. In addition, we calculate the error module rate~(EMR), which represents the error rate~(in percentage) of the modules in the QR Code.

\begin{figure}[htb]
    \centering
    \includegraphics[width=0.98\linewidth]{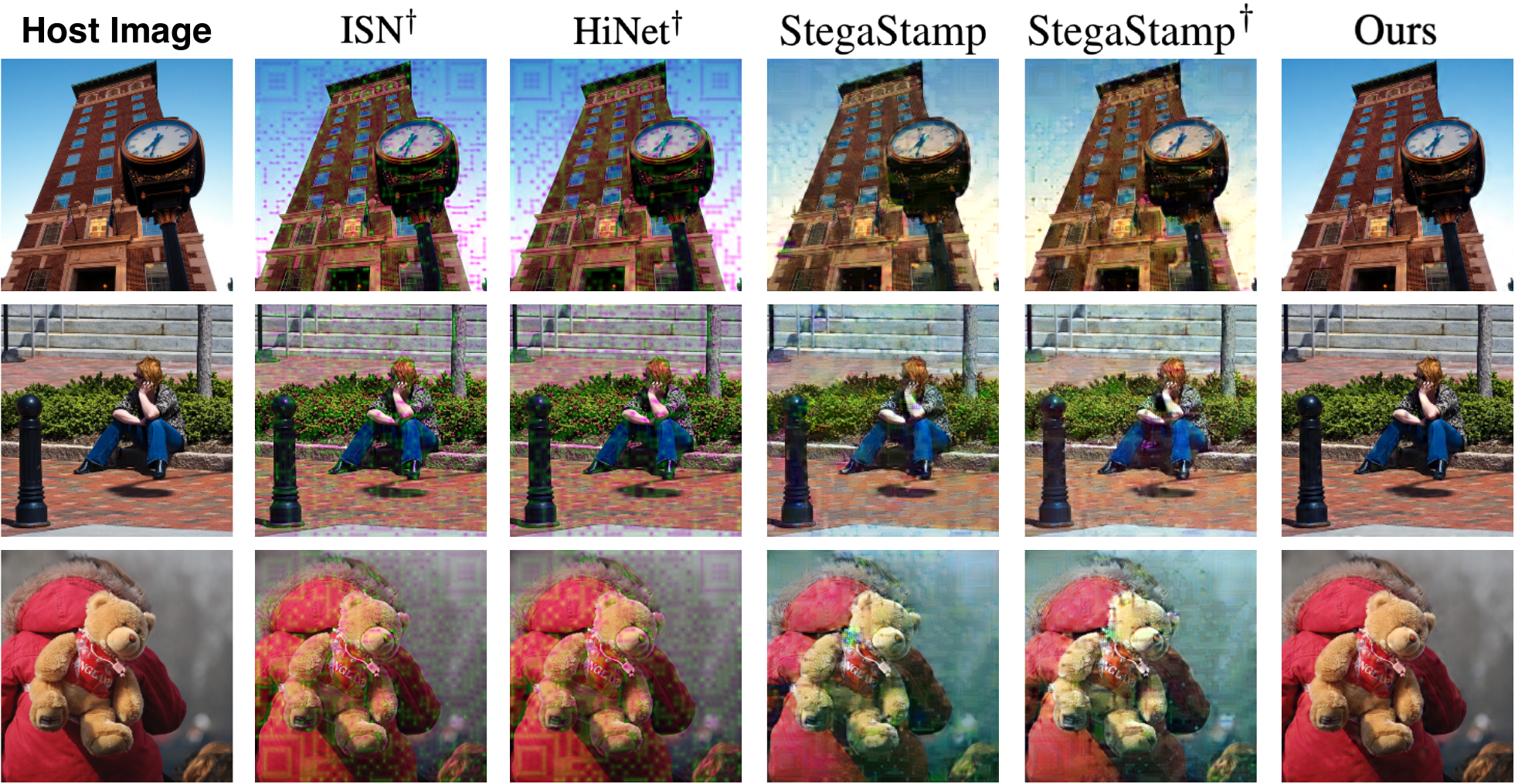}\vspace{-8pt}
    \caption{
        Stego images generated by different methods.
   }
  \label{fig: evl_quality_more} \vspace{-15pt}
\end{figure}

\textbf{Baselines} We compare our method with some state-of-the-art methods\footnote{Since RIIS~\cite{xu2022robust} has not released its source code or pre-trained model, we cannot compare with it.}, including ISN~\cite{lu2021large}, HiNet~\cite{jing2021hinet} and StegaStamp~\cite{tancik2020stegastamp}. Since these methods are not designed specifically for our task, we train these models on our datasets for a fair comparison. Moreover, for ISN and HiNet, since they are not robust steganography methods, we incorporate the distortion simulation module when training them. The re-trained models of these two methods are illustrated as ISN\textsuperscript{$\dagger$} and HiNet\textsuperscript{$\dagger$}, respectively. For StegaStamp, we also additionally train it by using the same distortion level as our method, represented as StegaStamp\textsuperscript{$\dagger$}.

\begin{figure*}[htb]
    \centering
    \includegraphics[width=1.0\linewidth]{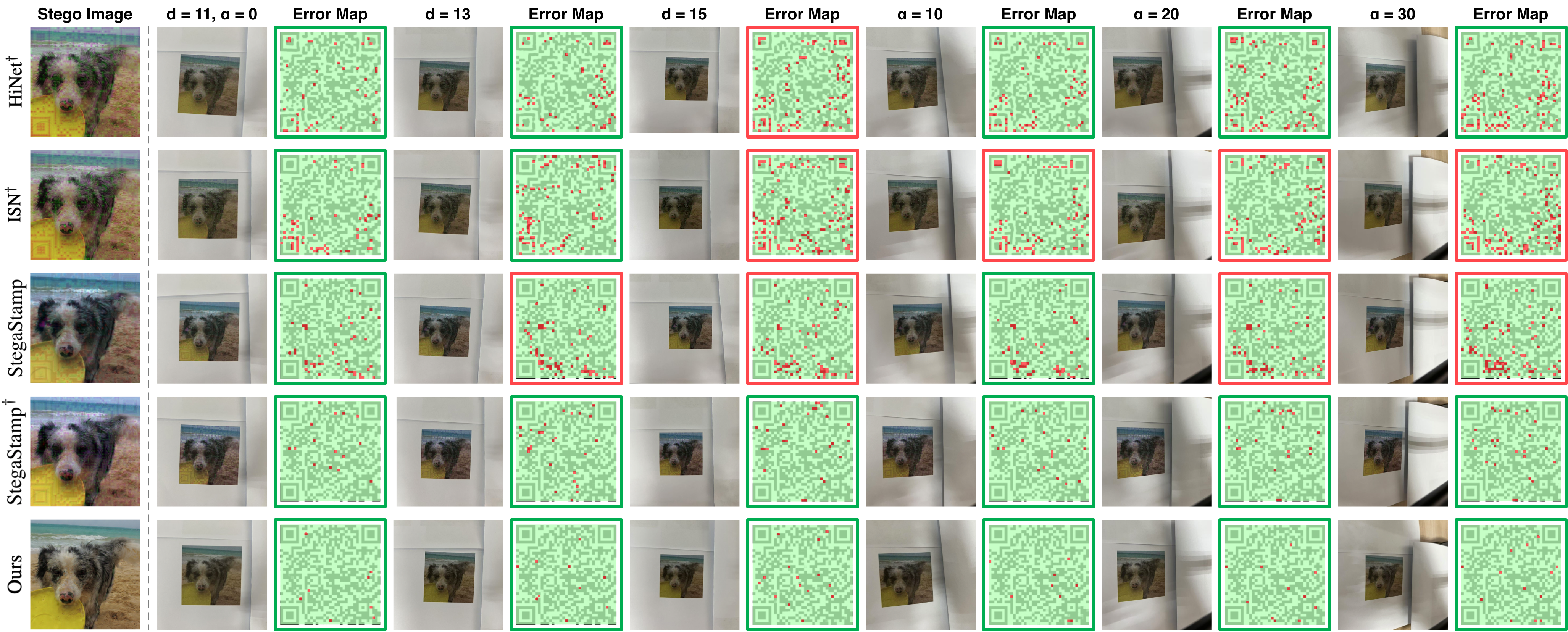}\vspace{-6pt}
    \caption{
        Photos of printed stego images and their decoding errors under different shooting situations. The actual image used for decoding is cropped out and resized from the photo. Photos shown in this figure are intended for an intuitive demonstration of the shooting results. QR Codes with green borders can be recognized while those with red borders cannot. Zoom in for better observation.
   }
  \label{fig: evl_shooting} 
  \vspace{-12pt}
\end{figure*}

\begin{figure}[t]
    \centering
    \includegraphics[width=0.98\linewidth]{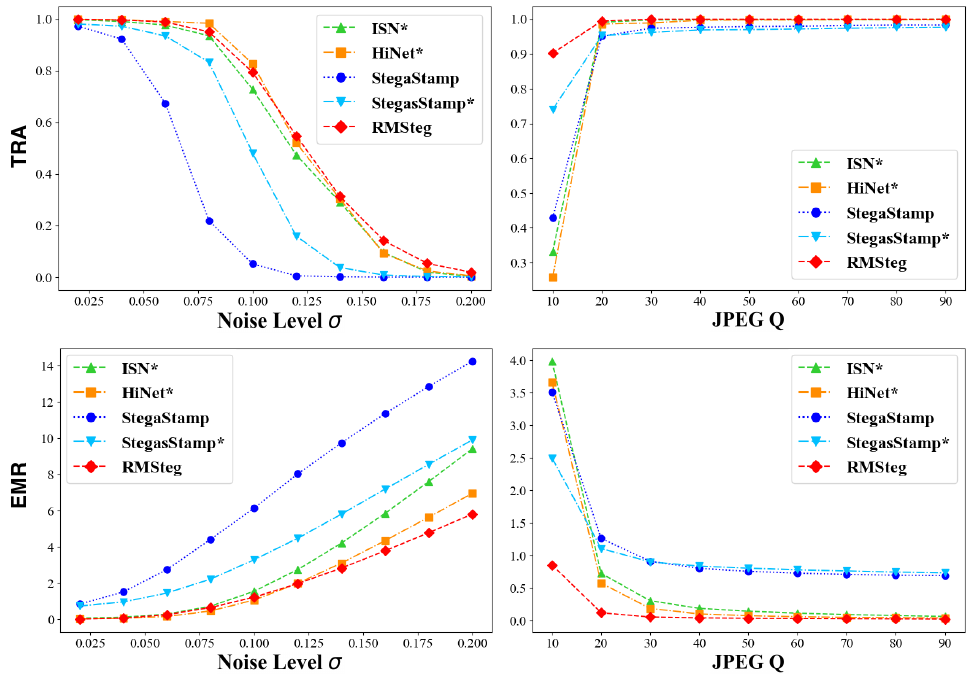}\vspace{-5pt}
    \caption{
        TRA and EMR under different levels of distortions.
   }
  \label{fig: evl_decoding} \vspace{-10pt}
  \vspace{-7pt}
\end{figure}

\subsection{Steganography Quality}
\label{sec: evl_quality}
Steganography quality indicates both stego image quality and decoding accuracy. To compare the robustness of different methods against image distortions, we first consider several manually created situations\footnote{The experiments under more situations are presented in the appendix.}: Gaussian noise, JPEG compression and random noise combinations. We generate random noise combinations~(represented by \textsl{Mixed}) with the distortion simulation module introduced in Sec.~\ref{sec: optim}. We then consider the printing and photography case to validate the methods' robustness against real-world distortions since it is one of the most extremely severe distortion situations and contains mixed disturbance factors. We randomly select 100 host images and embed random message in them. We then use a inkjet printer to print the encoded images out and take photos with a cell phone. To eliminate the potential errors caused by factors like print quality, we repeat the experiment for 5 times and choose the best result.

\autoref{tab: evl_steg_quality} shows the experiment results, \autoref{fig: evl_quality} and \autoref{fig: evl_quality_more} demonstrate some qualitative results. It can be observed that our method can achieve higher stego image quality, especially for LPIPS that represents the perceptual similarity. StegaStamp incorporates adversarial training~\cite{goodfellow2014generative} to make the generated stego image more realistic and it does work when facing low-level distortions. However, with the distortion used during training increases, StegaStamp\textsuperscript{$\dagger$} fails to preserve a sound visual quality and instead lead to hue shifting and artifacts. In addition, adversarial training can sometimes bring severe artifacts in some regions, as shown in the 3rd and 4th row of \autoref{fig: evl_quality}. For HiNet and ISN that both leverage normalizing flow, due to the lack of inner-channel interaction by using CNN-based affine block, the stego images they derive have obvious QR Code-like artifacts, making the existence of secret message easy to detect. In terms of decoding accuracy, although HiNet\textsuperscript{$\dagger$} outperforms our method in some cases, we achieve the best performance in the mixed noise and printing tests, which are more close to the real-world application scenarios.

\autoref{fig: evl_decoding} shows the decoding accuracy under more levels of distortions, which are Gaussian Noise whose standard deviation ranges from 0.02 to 0.2 and JPEG compression with quality ranging from 10 to 90. It can be observed that our method demonstrates a stable and good performance under these situations.

\begin{figure*}[t]
    \centering
    \includegraphics[width=1.0\linewidth]{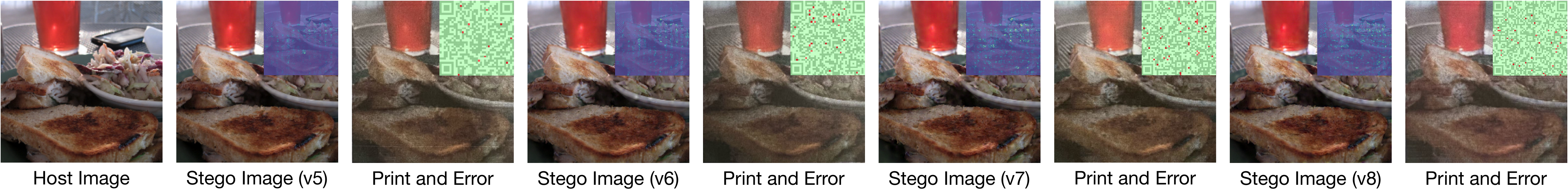}\vspace{-8pt}
    \caption{
        Stego images (the residual is shown in the upper right) and decoding results for printing situation with different code versions.
   }
  \label{fig: evl_capacity} \vspace{-6pt}
\end{figure*}

In practical application scenarios, the shooting condition may vary from time to time and a good message embedding method should keep its robustness in most cases. As a result, we further measure the decoding accuracy under different shooting situations. We mainly consider the shooting distance and angle~(the offset relative to vertical shooting). Given the default shooting distance and angle of this paper as 11\textsl{cm} and $0^{\circ}$, we gradually increase these two values and the test results are shown in \autoref{tab: evl_shooting}. It can be found that, with the shooting distance and angle grow, the EMRs exhibit a significant increase for all methods. However, for TRA that directly reflects the identifiability of QR Codes, our method maintains a fair performance. The comparison shown in \autoref{fig: evl_shooting} also indicates that RMSteg can achieve high decoding accuracy under different shooting situations.

\begin{table*}[htb]
    \caption{
        Decoding accuracy under different shooting situations. Here $d$ indicates the shooting distance~(measured by \textsl{cm}) and $\alpha$ is the shooting angle offset~(measured by \textsl{degree}). The best and second-best results are marked in \textcolor{red}{red} and \textcolor{blue}{blue} colors, respectively.
    }\vspace{-8pt}
    \newcolumntype{M}[1]{>{\centering\arraybackslash}m{#1}}
    \renewcommand\arraystretch{0.75}
    \centering
    \small
    \scalebox{1.02}{
    \begin{tabular}{M{1.8cm}M{0.8cm}M{0.8cm}M{0.8cm}M{0.8cm}M{0.8cm}M{0.8cm}M{0.8cm}M{0.8cm}M{0.8cm}M{0.8cm}M{0.8cm}M{0.8cm}} 
    \bottomrule
    \multirow{2}{=}{\centering{Method}} & \multicolumn{2}{c}{\scriptsize{$d = 11, \alpha = 0$}} & \multicolumn{2}{c}{\scriptsize{$d = 13$}} & \multicolumn{2}{c}{\scriptsize{$d = 15$}} & \multicolumn{2}{c}{\scriptsize{$\alpha = 10$}} & \multicolumn{2}{c}{\scriptsize{$\alpha = 20$}} & \multicolumn{2}{c}{\scriptsize{$\alpha = 30$}}\\ [-0.2pt]
     & \scriptsize{TRA$\uparrow$} & \scriptsize{EMR$\downarrow$} & \scriptsize{TRA$\uparrow$} & \scriptsize{EMR$\downarrow$} & \scriptsize{TRA$\uparrow$} & \scriptsize{EMR$\downarrow$} & \scriptsize{TRA$\uparrow$} & \scriptsize{EMR$\downarrow$} & \scriptsize{TRA$\uparrow$} & \scriptsize{EMR$\downarrow$} & \scriptsize{TRA$\uparrow$} & \scriptsize{EMR$\downarrow$} \\ [-0.2pt]
    \hline

    \scriptsize{~~ISN\textsuperscript{$\dagger$}} & \scriptsize{0.960} & \scriptsize{\sbest{1.125}} & \scriptsize{\sbest{0.920}} & \scriptsize{\sbest{2.173}} & \scriptsize{\sbest{0.790}} & \scriptsize{\sbest{2.780}} & \scriptsize{\sbest{0.910}} & \scriptsize{\sbest{1.859}} & \scriptsize{0.820} & \scriptsize{\sbest{2.394}} & \scriptsize{0.590} & \scriptsize{\sbest{3.389}} \\ [-0.2pt]

    \scriptsize{~~HiNet\textsuperscript{$\dagger$}} & \scriptsize{\sbest{0.970}} & \scriptsize{1.619} & \scriptsize{0.850} & \scriptsize{3.033} & \scriptsize{0.680} & \scriptsize{3.909} & \scriptsize{0.880} & \scriptsize{2.396} & \scriptsize{0.800} & \scriptsize{3.362} & \scriptsize{\sbest{0.660}} & \scriptsize{3.828} \\ [-0.2pt]

    \scriptsize{StegaStamp} & \scriptsize{0.750} & \scriptsize{3.214} & \scriptsize{0.610} & \scriptsize{4.131} & \scriptsize{0.350} & \scriptsize{5.321} & \scriptsize{0.540} & \scriptsize{3.784} & \scriptsize{0.360} & \scriptsize{4.258} & \scriptsize{0.040} & \scriptsize{5.916} \\ [-0.2pt]

    \scriptsize{StegaStamp\textsuperscript{$\dagger$}} & \scriptsize{0.900} & \scriptsize{1.917} & \scriptsize{0.790} & \scriptsize{2.469} & \scriptsize{0.710} & \scriptsize{2.922} & \scriptsize{0.860} & \scriptsize{2.272} & \scriptsize{\sbest{0.840}} & \scriptsize{2.413} & \scriptsize{0.570} & \scriptsize{3.978} \\ [-0.2pt]

    \scriptsize{~~Ours~~~} & \scriptsize{\best{1.000}} & \scriptsize{\best{0.606}} & \scriptsize{\best{1.000}} & \scriptsize{\best{0.891}} & \scriptsize{\best{0.980}} & \scriptsize{\best{1.269}} & \scriptsize{\best{1.000}} & \scriptsize{\best{0.953}} & \scriptsize{\best{1.000}} & \scriptsize{\best{1.163}} & \scriptsize{\best{0.960}} & \scriptsize{\best{1.680}} \\ [-0.2pt]

    \toprule
    \end{tabular}
    }
    \label{tab: evl_shooting}\vspace{-18pt}
\end{table*}

\begin{table}[htb]
    \caption{
        Model performance under different QR Code versions. The numbers in parentheses indicate the encoding capacity in \textsl{bit}.
    }\vspace{-6pt}
    \newcolumntype{M}[1]{>{\centering\arraybackslash}m{#1}}
    \renewcommand\arraystretch{0.8}
    \centering
    \small
    \scalebox{1.0}{
        \begin{tabular}{M{0.93cm}M{0.65cm}M{0.65cm}M{0.55cm}M{0.55cm}M{0.55cm}M{0.55cm}M{0.55cm}} 
        \bottomrule
        \multirow{2}{=}{\centering{\scriptsize{Version}}} & \multicolumn{3}{c}{\scriptsize{Stego Image}} & \multicolumn{2}{c}{\scriptsize{Mixed}} & \multicolumn{2}{c}{\scriptsize{Printing}} \\ [-0.2pt]
         & \scriptsize{PSNR$\uparrow$} & \scriptsize{SSIM$\uparrow$} & \scriptsize{LPIPS$\downarrow$} & \scriptsize{TRA$\uparrow$} & \scriptsize{EMR$\downarrow$} & \scriptsize{TRA$\uparrow$} & \scriptsize{EMR$\downarrow$} \\ [-0.2pt]
        \hline

        \scriptsize{v5~(368)} & \scriptsize{32.883} & \scriptsize{0.9109} & \scriptsize{0.0707} & \scriptsize{0.859} & \scriptsize{0.861} & \scriptsize{1.000} & \scriptsize{0.606} \\ [-0.2pt]

        \scriptsize{v6~(480)} & \scriptsize{31.363} & \scriptsize{0.8903} & \scriptsize{0.0892} & \scriptsize{0.859} & \scriptsize{0.934} & \scriptsize{1.000} & \scriptsize{0.877} \\ [-0.2pt]

        \scriptsize{v7~(528)} & \scriptsize{31.167} & \scriptsize{0.8880} & \scriptsize{0.0902} & \scriptsize{0.782} & \scriptsize{1.216} & \scriptsize{0.890} & \scriptsize{1.290} \\ [-0.2pt]

        \scriptsize{v8~(688)} & \scriptsize{30.765} & \scriptsize{0.8762} & \scriptsize{0.1020} & \scriptsize{0.743} & \scriptsize{1.370} & \scriptsize{0.820} & \scriptsize{1.476} \\ [-0.2pt]

        \toprule
        \end{tabular}
    }
    \label{tab: evl_capacity}\vspace{-14pt}
\end{table}

\subsection{Quality under Different Embedding Capacity}
\label{evl: capacity}

To validate the generality of our method, we test it under different embedding capacity, i.e., using QR Codes with different versions for training. We demonstrate the model performance on code versions from \textsl{v5} to \textsl{v8} in \autoref{tab: evl_capacity}. Although the steganography quality is getting worse with the embedding capacity increases, the artifacts in the stego images are still not perceptible, especially when the image is printed out, as shown in \autoref{fig: evl_capacity}. In addition, RMSteg keeps a TRA of more than 0.8 even for code \textsl{v8} whose embedding capacity is two more times higher than \textsl{v5}. Compared with StegaStamp~\cite{tancik2020stegastamp} that is also designed for robust message embedding, its PSNR is lower than 25 when encoding 200 bits in a $400 \times 400$ image according to the original paper. In contrast, our method is able to keep a PSNR of around 30 when encoding more than 600 bits in a $224 \times 224$ image. Thus, RMSteg can achieve a higher steganography quality and meanwhile the embedding capacity is much larger.

\subsection{Ablation Study}
\label{sec: ablation}
We conduct an ablation study to validate the effectiveness of the invertible QR Code transition~(IQRT), the invertible token fusion~(ITF) module and the AttnFlow model. The result is shown in \autoref{tab: evl_ablation}.

\textbf{IQRT} The model without IQRT performs slightly better than the full model in decoding accuracy. This is because IQRT may sometimes cause information loss, e.g., some code module could be wrongly transformed during this procedure, although the QR Code is still identifiable. On the other hand, IQRT can largely improve stego image quality.

\begin{figure}[]
    \centering
    \includegraphics[width=1.0\linewidth]{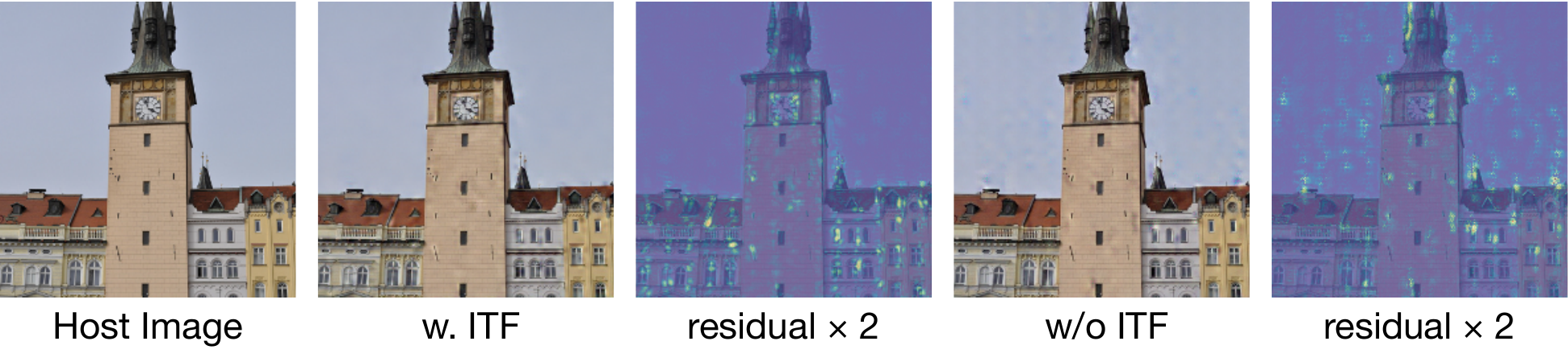}\vspace{-10pt}
    \caption{The generated stego images using and not using ITF.}
    \label{fig: cmp_itf}\vspace{-12pt}
\end{figure}

\textbf{ITF Module} As discussed in Sec.~\ref{sec: itf}, the ITF module can learn a transformation for image tokens and thus leading to better stego image quality. We also find that the ITF module can help derive a better distribution of artifacts brought by message embedding. As shown in \autoref{fig: cmp_itf}, the stego image generated using ITF has much less distortion in homogenous regions~(i.e., the sky), achieving a better visual quality.

\begin{table}[htb]
    \caption{
        Ablation study result. Here \textsl{CAT} indicates cross attention, \textsl{TR} represents tokenize representation. The best and second-best results are marked in \textcolor{red}{red} and \textcolor{blue}{blue} colors, respectively.
    }\vspace{-8pt}
    \newcolumntype{M}[1]{>{\centering\arraybackslash}m{#1}}
    \renewcommand\arraystretch{0.8}
    \centering
    \small
    \scalebox{1.02}{
        \begin{tabular}{M{1.8cm}M{0.72cm}M{0.72cm}M{0.72cm}M{0.72cm}M{0.72cm}} 
        \bottomrule
        \multirow{2}{=}{\centering{Method}} & \multicolumn{3}{c}{\scriptsize{Stego Image}} & \multicolumn{2}{c}{\scriptsize{Mixed}} \\ [-0.2pt]
        & \scriptsize{PSNR$\uparrow$} & \scriptsize{SSIM$\uparrow$} & \scriptsize{LPIPS$\downarrow$} & \scriptsize{TRA$\uparrow$} & \scriptsize{EMR$\downarrow$} \\ [-0.2pt]
        \hline

        \scriptsize{w/o IQRT} & \scriptsize{30.662} & \scriptsize{0.8651} & \scriptsize{0.1059} & \scriptsize{\best{0.871}} & \scriptsize{\best{0.828}} \\ [-0.2pt]

        \scriptsize{w/o ITF} & \scriptsize{31.422} & \scriptsize{0.8771} & \scriptsize{0.0919} & \scriptsize{0.833} & \scriptsize{0.947} \\ [-0.2pt]

        \scriptsize{w/o CAT} & \scriptsize{32.221} & \scriptsize{0.9036} & \scriptsize{0.0859} & \scriptsize{0.845} & \scriptsize{0.954} \\ [-0.2pt]

        \scriptsize{w/o TR + ISN} & \scriptsize{\sbest{32.444}} & \scriptsize{0.8856} & \scriptsize{0.3209} & \scriptsize{0.692} & \scriptsize{3.698} \\ [-0.2pt]

        \scriptsize{w/o TR + HiNet} & \scriptsize{31.513} & \scriptsize{0.8674} & \scriptsize{0.3370} & \scriptsize{0.714} & \scriptsize{3.321} \\ [-0.2pt]

        \scriptsize{1 AACB} & \scriptsize{30.426} & \scriptsize{0.8728} & \scriptsize{0.1076} & \scriptsize{0.798} & \scriptsize{1.135} \\ [-0.2pt]

        \scriptsize{2 AACBs} & \scriptsize{31.083} & \scriptsize{0.8972} & \scriptsize{0.0831} & \scriptsize{0.819} & \scriptsize{0.924} \\ [-0.2pt]

        \scriptsize{3 AACBs} & \scriptsize{31.649} & \scriptsize{\sbest{0.9008}} & \scriptsize{\sbest{0.0796}} & \scriptsize{0.819} & \scriptsize{0.924} \\ [-0.2pt]

        \scriptsize{Ours Full Model} & \scriptsize{\best{32.883}} & \scriptsize{\best{0.9109}} & \scriptsize{\best{0.0707}} & \scriptsize{\sbest{0.856}} & \scriptsize{\sbest{0.861}} \\ [-0.2pt]

        \toprule
        \end{tabular}
    }
    \label{tab: evl_ablation}\vspace{-8pt}
\end{table}

\textbf{Cross Attention in AACB} We train our model by removing the cross attention blocks defined in \autoref{eq: aacb_forward} and the result shows that this design can improve the overall model performance.

\textbf{Tokenized Representation} We replace the AttnFlow model with ISN and HiNet (two CNN-based normalizing flow model), respectively, to validate the effectiveness of introducing tokenized image representation (the IQRT module is retained). The result shows that, compared with CNN-based scheme, incorporating tokenized image representation makes normalizing flow more competent for robust steganography task.

\textbf{AACB Number} It shows that, increasing the number of AACBs can obtain a better model performance, which is consistent with the fundamental of normalizing flow.

%% file: sec/conclusion.tex
\section{Conclusion}

We propose a robust message embedding framework based on an attention flow-based model, called RMSteg. Our method is capable of generating stego images that can survive various real-world distortions, especially for printing and photography. To our best knowledge, RMSteg is the first method that introduces the transformer-based attention mechanism into normalizing flow. Our experiments show that this scheme is competent in steganography tasks. Compared with existing methods, RMSteg can achieve a better performance in robust and high-quality message embedding. We believe this is to a large extent due to the incorporation of the tokenized image representation and we hope this scheme can inspire subsequent studies.

%% file: sec/appendix.tex
\clearpage
\setcounter{page}{1}
\appendix
\maketitlesupplementary


\section{Details on Model Implementation}

\subsection{Invertible QR Code Transition}
We directly adopt the invertible neural network~(INN) architecture of ISN~\citesup{lu2021large_sup} to implement the QR Code transition procedure. Instead of the 16 invertible blocks used in the original paper, we only use 2 of them to lower the model complexity, since we empirically find that this transition does not need that large volume of parameters.

As mentioned in \cref{sec: qr_trans}, we employ a constraint on the transformed QR Code to ensure that it can be identified to restore the secret message. We adopt the same strategy as ArtCoder~\citesup{su2021artcoder_sup}, which is to simulate the most used \textsl{Google ZXing}~\citesup{zxing_2013_sup} rules that only read the center pixel of each module in a QR Code. According to Xu et al.~\citesup{xu2021art_sup}, the pixels closer to the module center should have a higher probability to be sampled. As a result, this sampling procedure can be modeled by performing a Gaussian convolution operation upon each code module. Specifically, given a QR Code with $n \times n$ modules of size $m \times m$, a Gaussian convolution kernel sized $m \times m$ is used to convolute each module with a stride of $m$ and derive an $n \times n$ sample result. The feature map is then binarized with a threshold, which is empirically set as 0.02 (given the pixel values range in $[0, 1]$) in this paper, since we find this threshold value can guarantee the identifiability of transformed QR Code. This means those pixels with values greater than 0.02 are regarded as white modules and the rest are black modules. During training, we obtain an error map $\xi$ which indicates the wrongly transformed code modules and backward the gradient for model optimization. This process is demonstrated in \cref{fig: sup_qr_conv} and the optimization target can be formulated as the following loss function:
\setcounter{equation}{0}
\renewcommand{\theequation}{S\arabic{equation}}
\renewcommand*{\theHequation}{\theequation}
\begin{equation}
    \begin{split}
    \mathcal{L}_t &= \left\| qc(I^{*}_q) \cdot \xi - qc(I_q) \cdot \xi \right\|_1, \\
    \xi &= \left\| bin_k(qc(I^{*}_q)) - qc(I_q) \right\|_1,
    \end{split}
\end{equation}
in which $I_q$ and $I^{*}_q$ represent the original QR Code and transformed result, respectively, $qc(\cdot)$ indicates the Gaussian kernel convolution and $bin_{k}(\cdot)$ is the binarize operation with the threshold as $k$. Since the aforementioned calculation is differentiable, it can be optimized jointly with other network modules during training. 

In our implementation, we resize the QR Code to a size of $5n \times 5n$ and use a $5 \times 5$ kernel to perform the convolution operation. The value of $n$ depends on the version of QR Code and it is 37 for version 5 that we adopt in this paper. For each subsequent version, the value of $n$ is 4 greater than the previous version, e.g., it is 41 for version 6 and so forth. \cref{fig: sup_qr_trans_result} shows some transition results, we also provide the QR Code and its error map after the aforementioned convolution and binarization operations. Although the transition can sometimes lead to some wrongly transformed code modules, it does not affect the identifiability.

\setcounter{figure}{0}
\renewcommand{\thefigure}{S\arabic{figure}}
\renewcommand*{\theHfigure}{\thefigure}
\begin{figure}[]
    \centering
    \includegraphics[width=0.98\linewidth]{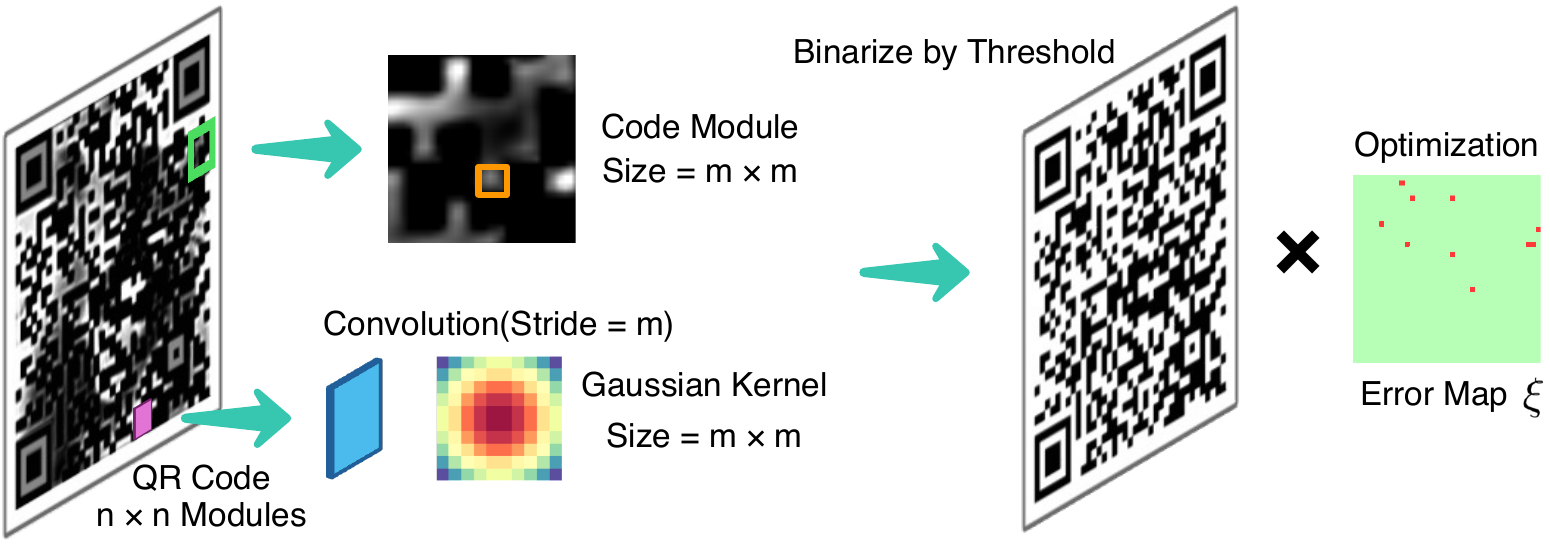}\vspace{-7pt}
    \caption{
        Demonstration of the QR Code scanning simulation.
   }
  \label{fig: sup_qr_conv}\vspace{-7pt}
\end{figure}

\begin{figure}[]
    \centering
    \includegraphics[width=0.98\linewidth]{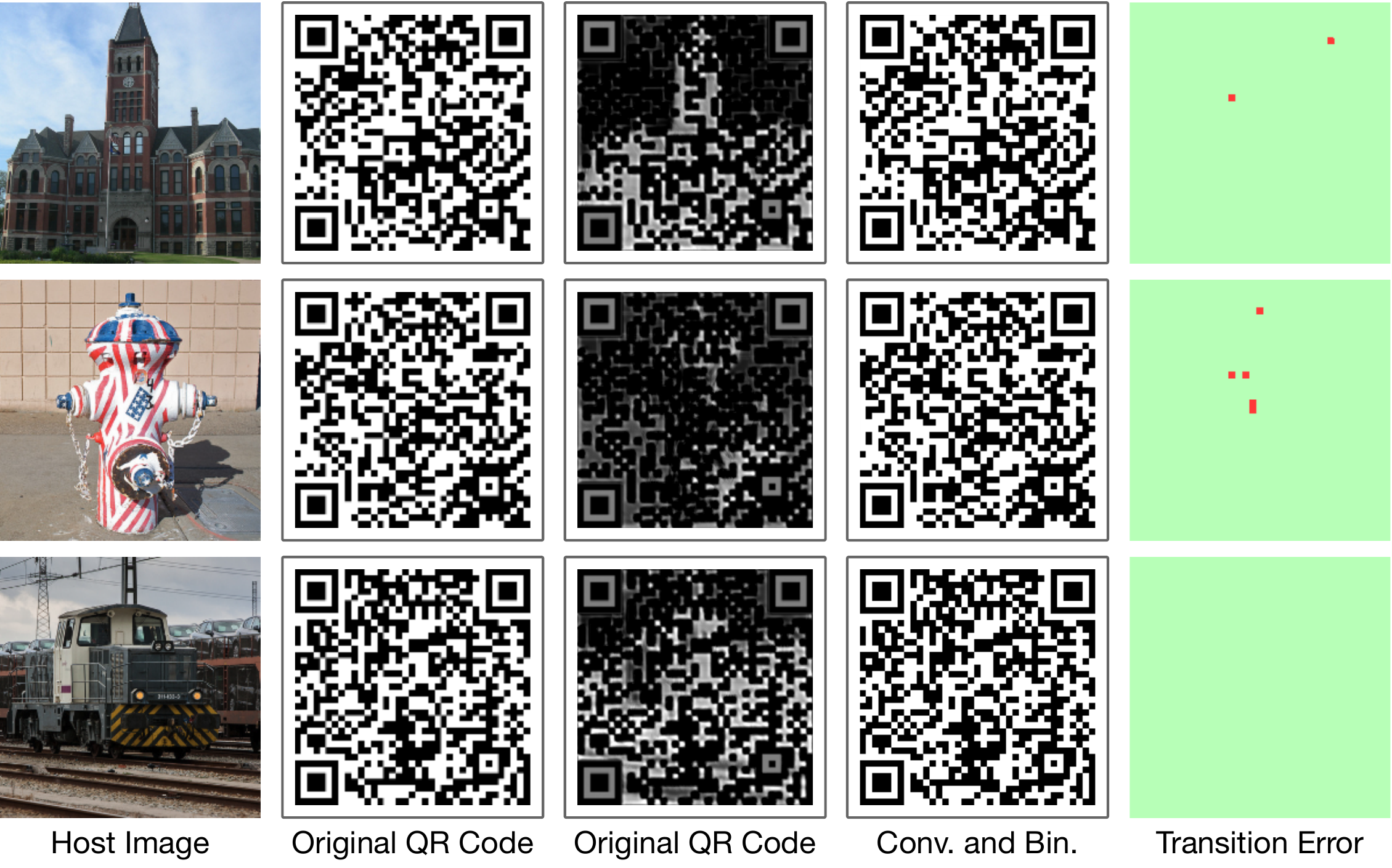}\vspace{-7pt}
    \caption{
        Some transition results. Here Conv. and Bin. indicate convolution and binarization, respectively.
   }
  \label{fig: sup_qr_trans_result}\vspace{-12pt}
\end{figure}

\subsection{AttnFlow Model}
We use some tokenizers to convert images to tokenized representations and some detokenizers to transform them back in the AttnFlow model. In our implementation, all the tokenizers used are based on the vision transformer~(ViT)~\citesup{dosovitskiy2020image_sup} architecture. We make some slight changes on the ViT-Base model that contains 12 transformer blocks with the token dimensionality and the multi-layer perceptron~(MLP) size being 768 and 3072, respectively. Specifically, we reduce the model complexity in our implementation by lowering the block number to 2 and the MLP size to 2048. The patch size used is $16 \times 16$. For the detokenizer, we simply use an MLP for dimension projection followed by a reshaping operation and two convolutional layers with GELU~\citesup{hendrycks2016gaussian_sup} as activation function to convert the tokens back to image.

For the self-attnetion and cross-attnetion blocks used in AACB, they have the same token dimensionality and MLP size as the tokenizers. We choose to use 4 AACBs in our full model since we find this block number makes a good balance between model performance and training cost.

\section{Experiment Details}

\subsection{Metrics Calculation}

\textbf{LPIPS} We use LPIPS~\citesup{zhang2018unreasonable_sup} as part of the optimization target during model training and one of the metrics for stego image quality evaluation. We calculated it with a VGG~\citesup{DBLP:journals/corr/SimonyanZ14a_sup} model pre-trained on ImageNet~\citesup{deng2009imagenet_sup}.

\textbf{TRA} We calculate the text recovery accuracy~(TRA) with the following scheme:
\begin{equation}
    TRA(\hat{I}_{qr}) = 
    \begin{cases}
        1.0 & \text{if}~\hat{I}_{qr}~\text{is identifiable} \\
        0.0 & \text{otherwise}
    \end{cases},
\end{equation}
where $\hat{I}_{qr}$ indicates the decoded QR Code. The final TRA is average value upon the whole testing dataset.

\textbf{EMR} We calculate the error module rate~(EMR) by measuring the wrongly decoded QR Code modules. Given a decoded QR Code, we binarize and compare it with the ground truth to derive the error rate.

\begin{figure}[]
    \centering
    \includegraphics[width=1.0\linewidth]{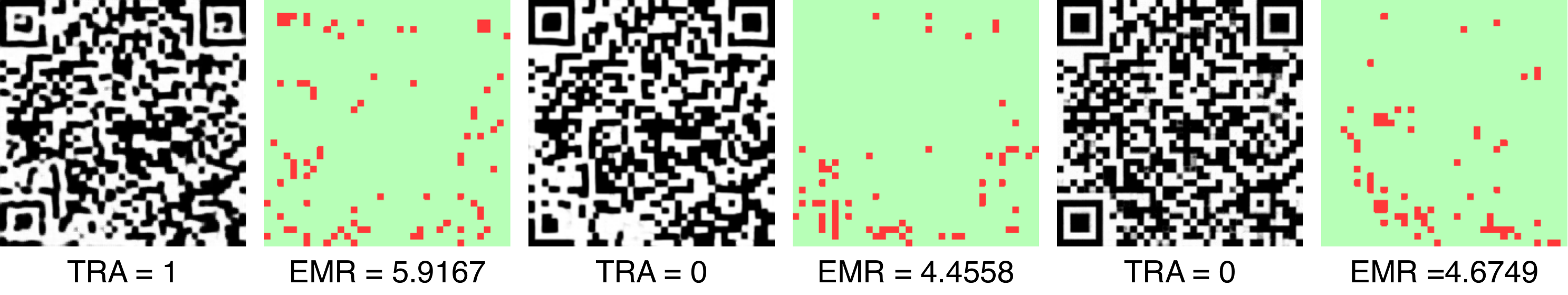}\vspace{-5pt}
    \caption{
        An example of decoded QR Code with higher EMR but lower TRA.
   }
  \label{fig: sup_emr_tra}
\end{figure}

It is worth noticing that, although low EMR can generally guarantee a high TRA, these two metrics are not necessarily positively correlated. An example is shown in \cref{fig: sup_emr_tra}, the second and third decoded QR Codes have lower EMR than the first one but they are not identifiable. This can be caused by two reasons, the first is that the finder and alignment patterns are damaged, making the code cannot be detected, corresponding to the second case in \cref{fig: sup_emr_tra}~(the finder pattern in the lower left is damaged). The second is that, a high error rate is caused in a small area, making the error correction~(ECC) scheme of QR Code fail to restore the error, corresponding to the third case in \cref{fig: sup_emr_tra}.

\subsection{Training Details}
Our model is implemented with PyTorch~\citesup{paszke2019pytorch_sup} and is trained on 4 NVIDIA GeForce 3090 GPUs. We set the batch size as 8 for each GPU and the mode is trained for 50K iterations. The initial learning rate is 0.0001, which decays by $10\%$ after each epoch until it reaches 0.00001. The model is optimized with the AdamW optimizer~\citesup{loshchilov2017decoupled_sup} with $\beta_1 = 0.9$ and $\beta_2 = 0.999$. The hyper-parameters in the loss function~(\autoref{eq: loss}) are set as $\alpha = 5.0$, $\beta = 0.2$, $\gamma = 3.5$, $\delta = 16$ and $\epsilon = 3.0$. The trainnable parameters $\alpha_i$ introduced in \autoref{eq: aacb_forward} are initialized as 0.01 in the beginning. The training host images are randomly cropped from the original images as 224 $\times$ 224 patches. The overall training process takes for about 13 hours.

\subsection{Distortion Simulation}
We adopt the same types of distortions as StegaStamp~\citesup{tancik2020stegastamp_sup} with some changes in the hyper-parameters. We provide a comparison of the settings of StegaStamp and our in \cref{tab: sup_distortion}. We largely increase the Gaussian noise level to fit for our task. We lower the distortion level of JPEG compression since we find that there is no need to use a very low JPEG compression quality for training to achieve sufficient robustness when a high-level Gaussian noise is employed. For the parameter of transition, since we manually crop the stego image out from the photo, rather than using object detection as StegaStamp, we do not need such a high parameter setting to promise enough robustness. As for how the distortion simulation is implemented and how do these parameters work, we strongly suggest referring to the description in the original paper~\citesup{tancik2020stegastamp_sup}.

\setcounter{table}{0}
\renewcommand{\thetable}{S\arabic{table}}
\renewcommand*{\theHtable}{\thetable}
\begin{table}[]
    \caption{
        Comparison of distortion parameters used by StegaStamp and RMSteg. Here \textsl{Bri.} is brightness, \textsl{Sat.} is saturation, \textsl{Noi.} is Gaussian noise level and \textsl{Tra.} is transition. 
    }\vspace{-7pt}
    \newcolumntype{M}[1]{>{\centering\arraybackslash}m{#1}}
    \renewcommand\arraystretch{1.0}
    \centering
    \small
    \scalebox{0.963}{
    \begin{tabular}{M{1.15cm}M{0.35cm}M{0.42cm}M{0.3cm}M{1.0cm}M{0.4cm}M{0.3cm}M{0.43cm}M{0.4cm}} 
    \bottomrule
    \scriptsize{Method} & \scriptsize{Bri.~~~} & \scriptsize{Hue~~~} & \scriptsize{Sat.} & \scriptsize{Contrast} & \scriptsize{jpeg} & \scriptsize{Noi.} & \scriptsize{Blur} & \scriptsize{~Tra.~~~} \\ [-0.2pt]
    \hline
    \scriptsize{StegaStamp} & \scriptsize{0.3} & \scriptsize{0.1} & \scriptsize{1.0} & \scriptsize{$[0.5, 1.5]$} & \scriptsize{25} & \scriptsize{0.02} & \scriptsize{7} & \scriptsize{0.10} \\ [-0.2pt]
    \scriptsize{Ours} & \scriptsize{0.3} & \scriptsize{0.1} & \scriptsize{1.0} & \scriptsize{$[0.5, 1.5]$} & \scriptsize{\textbf{60}} & \scriptsize{\textbf{0.07}} & \scriptsize{7} & \scriptsize{\textbf{0.02}} \\ [-0.2pt]
    \toprule
    \end{tabular}
    }
    \label{tab: sup_distortion}\vspace{-6pt}
\end{table}

\begin{figure}[]
    \centering
    \includegraphics[width=1.0\linewidth]{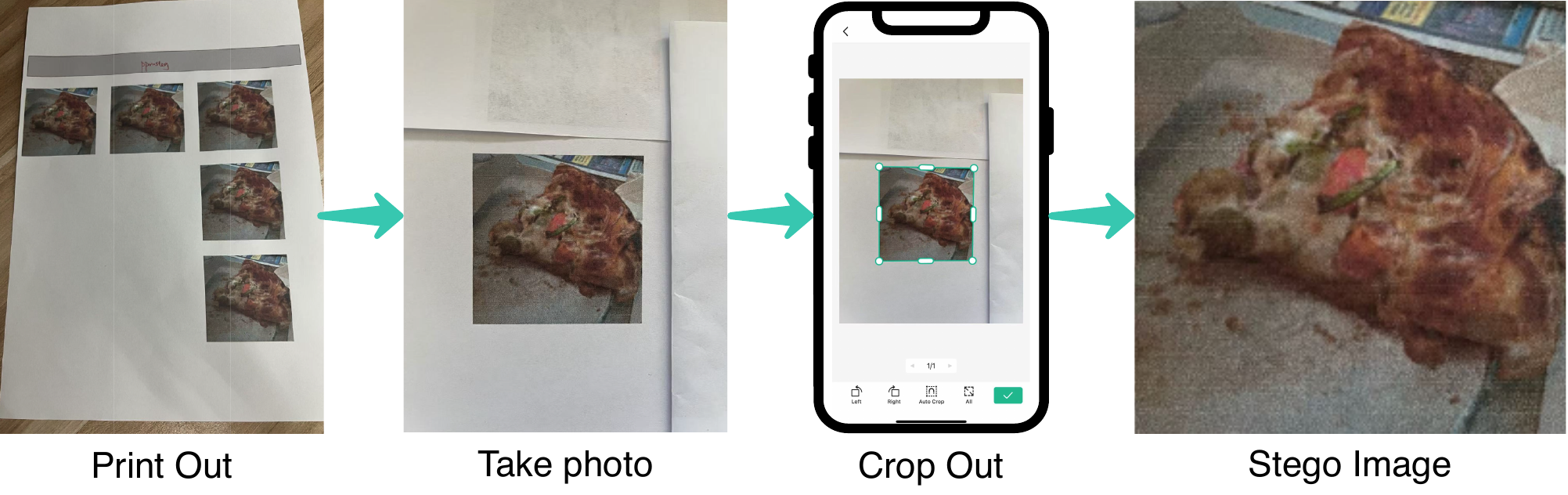}\vspace{-5pt}
    \caption{
        How we derive the stego image in the printing test.
   }
  \label{fig: sup_print_workflow}
\end{figure}

\subsection{Printing and Photography}
In this paper, we choose to use the case of printing and photography to measure the method robustness under extreme real-world distortions. During experiment, we first print the stego images out and take photos for them. For one stego image, we print it for multiple times on the same paper to prevent the fluctuation of printer's printing quality. Then, we manually crop the stego images out from the photos using CamScanner~\citesup{camscanner_sup} and use them for decoding test. A demonstration of the above workflow is shown in \cref{fig: sup_print_workflow}.

The printer we used for experiment is an HP OfficeJet Pro 8710 inkjet printer. We choose the printing quality of `Normal'~(among `Draft', `Normal' and `High'). The printed image size on paper is 5.4\textsl{cm} $\times$ 5.4\textsl{cm}. We take photos with an iPhone 13 Pro indoors under regular illumination. We take 5 photos for each image and choose the best value for final metrics calculation. For the experiments under different shooting angles, we maintain a vertical distance of 11~\textsl{cm} between the lens and the paper, which is our default shooting distance.

\section{Further Experiment and Discussion}

\subsection{Trainnable Coefficients in AACB}

We set the coefficient $\alpha_i$ of the cross-attention item in the AACB transformation function~(\autoref{eq: aacb_forward}) as a trainnable parameter to let the network learn by itself. We initialize this parameter as 0.01 at the beginning and optimize it during training. Here we provide the final converged values of $\alpha_i$ in the models with different numbers of AACBs. The result is shown in \cref{tab: sup_aacb_alpha}.

\begin{table}[]
    \caption{
        The $\alpha_i$ values when using different numbers of AACBs. 
    }\vspace{-7pt}
    \newcolumntype{M}[1]{>{\centering\arraybackslash}m{#1}}
    \renewcommand\arraystretch{1.0}
    \centering
    \small
    \scalebox{1.0}{
    \begin{tabular}{M{1.5cm}M{1.18cm}M{1.18cm}M{1.18cm}M{1.18cm}} 
    \bottomrule
    \scriptsize{Block Number} & \scriptsize{$\alpha_1$} & \scriptsize{$\alpha_2$} & \scriptsize{$\alpha_3$} & \scriptsize{$\alpha_4$} \\ [0.8pt]
    \hline
    \scriptsize{1} & \scriptsize{0.1354} & \scriptsize{--} & \scriptsize{--} & \scriptsize{--} \\ [-2.0pt]
    \scriptsize{2} & \scriptsize{0.0062} & \scriptsize{0.1425} & \scriptsize{--} & \scriptsize{--} \\ [-2.0pt]
    \scriptsize{3} & \scriptsize{0.0070} & \scriptsize{0.0915} & \scriptsize{0.0087} & \scriptsize{--} \\ [-2.0pt]
    \scriptsize{4} & \scriptsize{0.0022} & \scriptsize{0.1350} & \scriptsize{0.0745} & \scriptsize{0.0070} \\ [-0.8pt]
    \toprule
    \end{tabular}
    }
    \label{tab: sup_aacb_alpha}\vspace{-10pt}
\end{table}

\subsection{Anti-Distortion Ability}
\label{sec: sup_anti_distortion}
We evaluate the anti-distortion ability of our method in the paper. Here we further consider more real-world image distortions.

\begin{figure}[]
    \centering
    \includegraphics[width=1.0\linewidth]{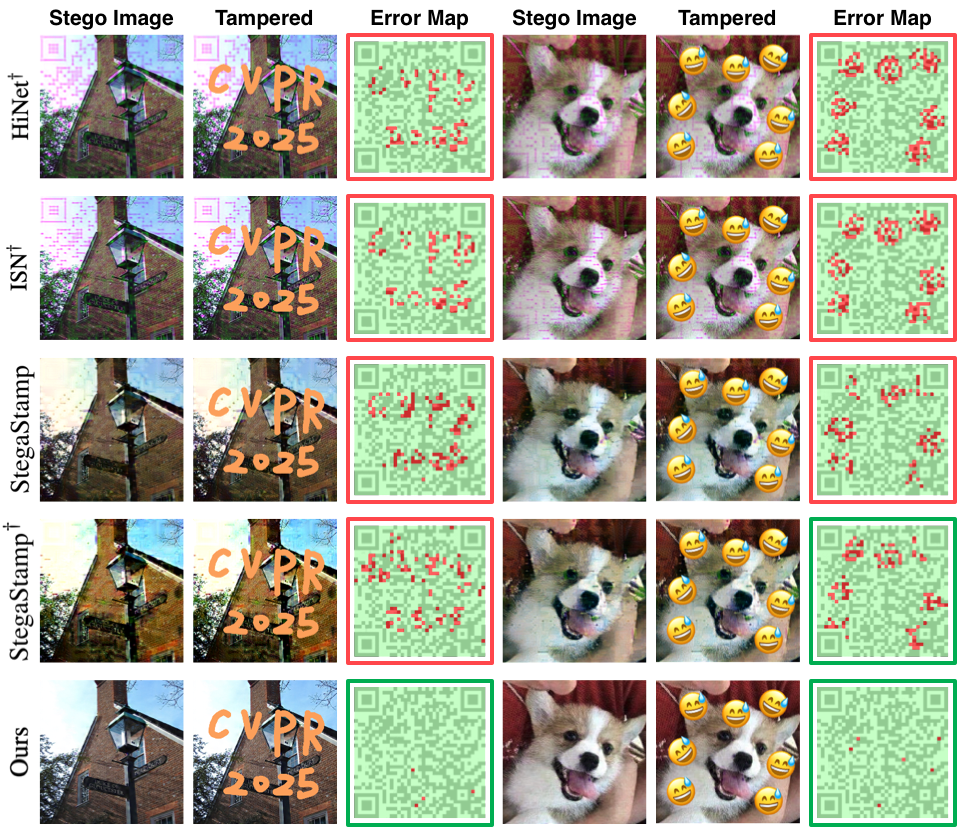}\vspace{-5pt}
    \caption{
        Decoding results under image tampering. QR Codes with green borders can be successfully identified while those with red borders cannot. Zoom in for better observation.
   }
  \label{fig: sup_tamper}\vspace{-10pt}
\end{figure}

\textbf{Tampering} During image transmission, tampering is one of the most common and severest distortions. We randomly tamper~(in our implementation, we use some black squares and mask them on the stego images) a certain ratio of the areas in the stego image and calculate the decoding accuracy to measure the robustness against this kind of distortion. The results are shown in \cref{tab: sup_anti_tamper}. It can be observed that, our method has significantly better robustness against tampering than previous methods. We speculate that, after introducing the transformer architecture into normalizing flow, secret message can be embedded into host images in a manner similar to redundant coding due to the inner-channel feature interaction brought by attention mechanism. This allows for the correct recovery of information even when some areas of the image are tampered. Take the two cases shown in \cref{fig: sup_tamper} as example, our method can achieve a high decoding accuracy under both cases. However, the remaining four CNN-based methods have high error rates in the tampered regions. This can to some extent prove that, CNN-based method tends to hide secret message in corresponding areas. Therefore, when a certain area is damaged, the corresponding area of the secret message will also fail to decode correctly. More results are provided in \cref{sec: sup_addition}.

\begin{table*}[]
    \caption{
        Decoding accuracy under tampering rate $r$. The best and second-best results are marked in \textcolor{red}{red} and \textcolor{blue}{blue} colors.
    }\vspace{-8pt}
    \newcolumntype{M}[1]{>{\centering\arraybackslash}m{#1}}
    \renewcommand\arraystretch{0.75}
    \centering
    \small
    \scalebox{1.0}{
    \begin{tabular}{M{1.24cm}M{0.7cm}M{0.7cm}M{0.7cm}M{0.7cm}M{0.7cm}M{0.7cm}M{0.7cm}M{0.7cm}M{0.7cm}M{0.7cm}M{0.7cm}M{0.7cm}M{0.7cm}M{0.7cm}} 
    \bottomrule
    \multirow{2}{=}{\centering{Method}} & \multicolumn{2}{c}{\scriptsize{$r = 5\%$}} & \multicolumn{2}{c}{\scriptsize{$r = 10\%$}} & \multicolumn{2}{c}{\scriptsize{$r = 15\%$}} & \multicolumn{2}{c}{\scriptsize{$r = 20\%$}} & \multicolumn{2}{c}{\scriptsize{$r = 25\%$}} & \multicolumn{2}{c}{\scriptsize{$r = 30\%$}} & \multicolumn{2}{c}{\scriptsize{$r = 35\%$}}\\ [-0.2pt]
     & \scriptsize{TRA$\uparrow$} & \scriptsize{EMR$\downarrow$} & \scriptsize{TRA$\uparrow$} & \scriptsize{EMR$\downarrow$} & \scriptsize{TRA$\uparrow$} & \scriptsize{EMR$\downarrow$} & \scriptsize{TRA$\uparrow$} & \scriptsize{EMR$\downarrow$} & \scriptsize{TRA$\uparrow$} & \scriptsize{EMR$\downarrow$} & \scriptsize{TRA$\uparrow$} & \scriptsize{EMR$\downarrow$} & \scriptsize{TRA$\uparrow$} & \scriptsize{EMR$\downarrow$} \\ [-0.2pt]
    \hline

    \scriptsize{~~ISN\textsuperscript{$\dagger$}} & \scriptsize{0.571} & \scriptsize{2.878} & \scriptsize{0.349} & \scriptsize{5.008} & \scriptsize{0.110} & \scriptsize{7.137} & \scriptsize{0.005} & \scriptsize{9.259} & \scriptsize{0.000} & \scriptsize{11.39} & \scriptsize{0.000} & \scriptsize{14.24} & \scriptsize{0.000} & \scriptsize{16.37} \\ [-0.2pt]

    \scriptsize{~~HiNet\textsuperscript{$\dagger$}} & \scriptsize{0.575} & \scriptsize{\sbest{2.698}} & \scriptsize{0.354} & \scriptsize{\sbest{4.725}} & \scriptsize{0.105} & \scriptsize{6.723} & \scriptsize{0.007} & \scriptsize{8.736} & \scriptsize{0.000} & \scriptsize{10.78} & \scriptsize{0.000} & \scriptsize{13.48} & \scriptsize{0.000} & \scriptsize{15.53} \\ [-0.2pt]

    \scriptsize{StegaStamp} & \scriptsize{0.915} & \scriptsize{3.221} & \scriptsize{0.673} & \scriptsize{5.080} & \scriptsize{0.245} & \scriptsize{6.902} & \scriptsize{0.029} & \scriptsize{8.716} & \scriptsize{0.001} & \scriptsize{10.54} & \scriptsize{0.000} & \scriptsize{12.89} & \scriptsize{0.000} & \scriptsize{14.68} \\ [-0.2pt]

    \scriptsize{StegaStamp\textsuperscript{$\dagger$}} & \scriptsize{\sbest{0.918}} & \scriptsize{3.038} & \scriptsize{\sbest{0.732}} & \scriptsize{4.801} & \scriptsize{\sbest{0.336}} & \scriptsize{\sbest{6.539}} & \scriptsize{\sbest{0.050}} & \scriptsize{\sbest{8.277}} & \scriptsize{\sbest{0.001}} & \scriptsize{\sbest{9.994}} & \scriptsize{\sbest{0.000}} & \scriptsize{\sbest{12.30}} & \scriptsize{\sbest{0.000}} & \scriptsize{\sbest{14.05}} \\ [-0.2pt]

    \scriptsize{~~Ours~~~} & \scriptsize{\best{0.996}} & \scriptsize{\best{0.119}} & \scriptsize{\best{0.995}} & \scriptsize{\best{0.340}} & \scriptsize{\best{0.966}} & \scriptsize{\best{0.755}} & \scriptsize{\best{0.818}} & \scriptsize{\best{1.324}} & \scriptsize{\best{0.533}} & \scriptsize{\best{2.059}} & \scriptsize{\best{0.233}} & \scriptsize{\best{3.192}} & \scriptsize{\best{0.113}} & \scriptsize{\best{4.195}} \\ [-0.2pt]

    \toprule
    \end{tabular}
    }
    \label{tab: sup_anti_tamper}\vspace{-6pt}
\end{table*}

\begin{figure*}[]
    \centering
    \includegraphics[width=1.0\linewidth]{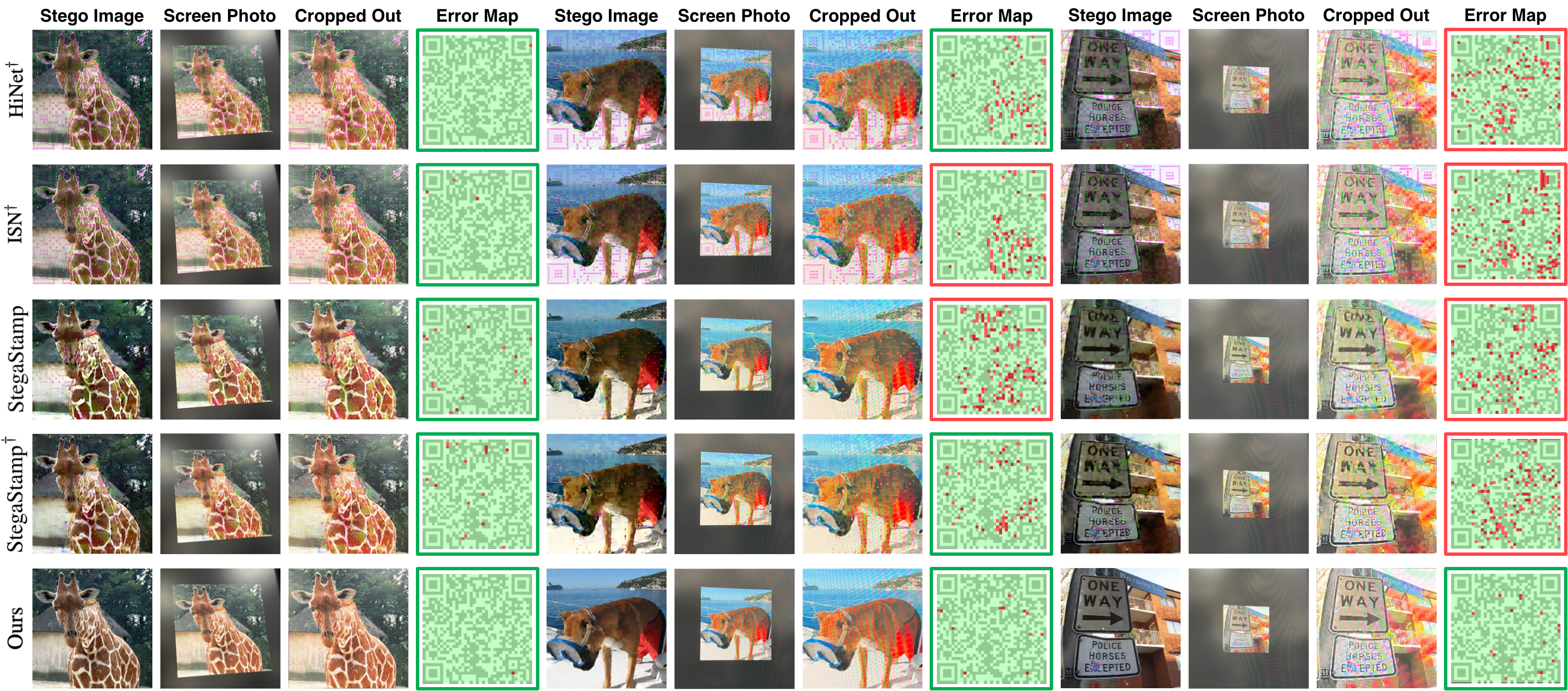}\vspace{-5pt}
    \caption{
        Stego images and decoding results under the distortion of light field messaging. QR Codes with green borders can be successfully identified while those with red borders cannot. Zoom in for better observation.
   }
  \label{fig: sup_lfm}\vspace{-8pt}
\end{figure*}

\textbf{Light Field Messaging} Wengrowski et al.~\citesup{wengrowski2019light_sup} consider the message embedding robustness against on-screen shooting, e.g., taking photo of the stego image displayed on a PC screen. Since the quantitative study upon this topic requires a large amount of experiments, such as the influence of different displayers and camera lenses, which are not the main contribution of this paper, here we just provide some qualitative results. We use an iPhone 13 Pro as the shooting camera and the displayer used to show the stego images is a BenQ~EW2770QZ with 2560$\times$1440 resolution. As shown in \cref{fig: sup_lfm}, we choose different shooting distances for a comprehensive demonstration. It can be observed that, with the shooting distance grows, the distortion of Moiré pattern becomes more and more obvious. However, compared with other methods, our method can keep a high decoding accuracy against this kind of distortion. We believe this can again prove the superiority of transformer-based scheme in handling robust steganography task. More results are provided in \cref{sec: sup_addition}.

\subsection{Ablation Study}
Since the quantitative results have been provided in \cref{sec: ablation}, here we mainly focus on the ablation experiment implementation details and qualitative comparison results.

\begin{figure*}[]
    \centering
    \includegraphics[width=1.0\linewidth]{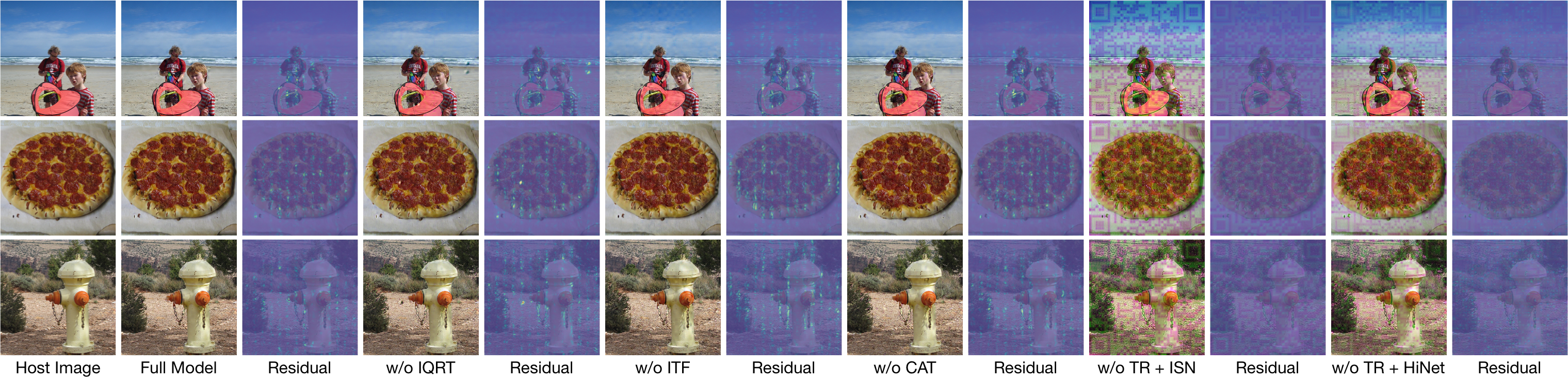}\vspace{-5pt}
    \caption{
        Qualitative results of the ablation study. Zoom in for better observation.
   }
  \label{fig: sup_ablation}
\end{figure*}

\textbf{IQRT} We validate the effectiveness of invertible QR Code transition by removing it from the pipeline, which means we directly tokenize the QR Code image and feed the tokens to the ITF module and then to the AttnFlow model. As shown in \cref{fig: sup_ablation}, the stego images generated without IQRT have more artifacts. As discussed in the paper, since we only apply one constraint on the transformed QR Code to guarantee its identifiability, the learned transition strategy will tend to help achieve a better steganography quality. However, as shown in \cref{tab: evl_ablation}, IQRT can lead to a slightly worse decoding accuracy. This is due to the information loss that sometimes happens during the transition process. Two examples can be found in \cref{fig: sup_qr_trans_result}. Overall, we believe this module can effectively improve the stego image quality. 

\textbf{ITF Module} The invertible token fusion module learns a transform matrix for QR Code image tokens. Compared with the invertible 1$\times$1 convolution~(IConv) proposed by GLOW~\citesup{kingma2018glow_sup}, our ITF learns a patch-wise~(or, token-wise) transformation instead of a channel-wise one. We believe it can rearrange the tokens just like IConv that can re-permute the channels in order to compensate for the insufficient distribution transformation ability of normalizing flow due to the affine-formed functions that have to be adopted for the invertibility of the model. The experiment result proves the competence of ITF module and as introduced in the paper, we empirically find that this module can help derive a better distribution of the artifacts in the stego image. Some results are also shown in \cref{fig: sup_ablation}, we believe this is to some extent due to the token rearranging brought by ITF.

\textbf{Cross Attention in AACB} We introduce an extra cross-attention item in \autoref{eq: aacb_forward}. We make this design because we hope more feature interactions can happen between host image tokens and QR Code tokens. As a result, we incorporate the cross-attention mechanism and add it to the affine transformation function. As shown in \cref{fig: sup_ablation}, this module can improve the visual quality of the stego image. The results in \cref{tab: evl_ablation} also shows that, the model with cross-attention has an around $15\%$ improvement in LPIPS.

\textbf{Tokenized Representation} We incorporate tokenized representation~(TR) in ITF module and AttnFlow model. To validate the effectiveness of TR, we remove the ITF module and replace the AttnFlow with CNN-based normalizing flow models~(ISN~\citesup{lu2021large_sup} and HiNet~\citesup{jing2021hinet_sup}). As the results shown in \cref{fig: sup_ablation}, the stego images generated are similar to that of original ISN and HiNet, containing obvious QR Code-like artifacts. This proves that CNN-based normalizing flow struggles to achieve a high-quality feature learning in the robust steganography task. In contrast, our transformer-based scheme extends the model ability and can help generate stego images with high visual similarity.

\subsection{Limitation Analysis and Future Work}

\begin{figure}[]
    \centering
    \includegraphics[width=1.0\linewidth]{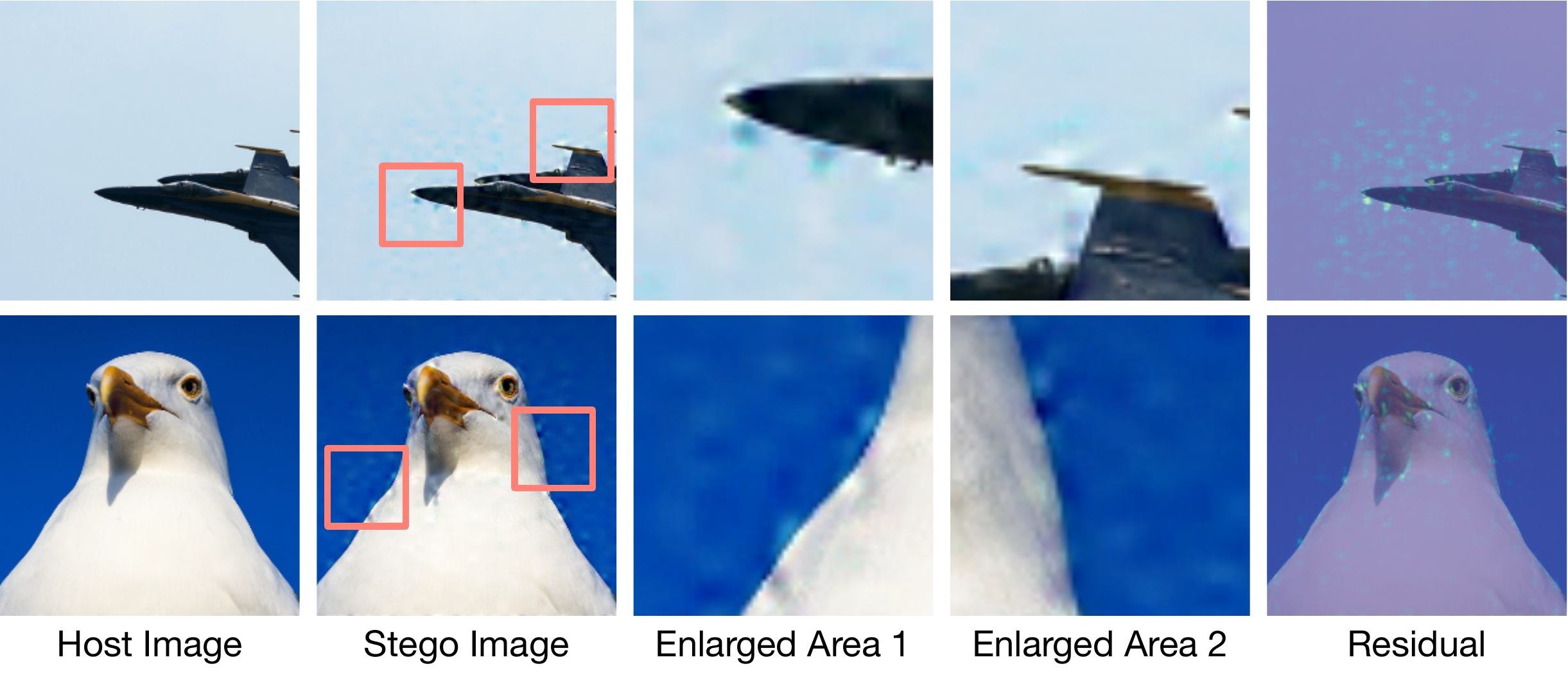}\vspace{-5pt}
    \caption{
        Generated stego images when facing host images with many homogeneous areas.
   }
  \label{fig: sup_homogeneous}
\end{figure}

Although our RMSteg can achieve the state-of-the-art performance in robust message embedding, it still has some limitations currently. First, as mentioned in \cref{sec: ablation}, our method can help distribute the steganography residual in heterogeneous regions to avoid perceptible artifacts. However, when facing host images with many homogeneous regions, our method can fail to preserve a good visual quality. It can be observed from \cref{fig: sup_homogeneous} that, although the artifacts is mainly concentrated in heterogeneous areas, the ratio of this kind of regions is too small to preserve the overall visual quality. Secondly, although the embedding capacity of RMSteg far exceeds previous methods, it still cannot conceal large-scale secret information, e.g., multiple images. 

In summary, we are going to focus on the two aforementioned limitations, i.e., better steganography quality and higher embedding capacity, in the future. We will explore more schemes to extend the performance of transformer-based steganography method. In addition, as mentioned in \cref{sec: sup_anti_distortion}, we will consider more kind of real-world image distortions to improve the method's applicability.

\subsection{Additional Results}
\label{sec: sup_addition}
In this section, we provide more qualitative results of the experiments mentioned in the paper and appendix.

\begin{figure*}[]
    \centering
    \includegraphics[width=1.0\linewidth]{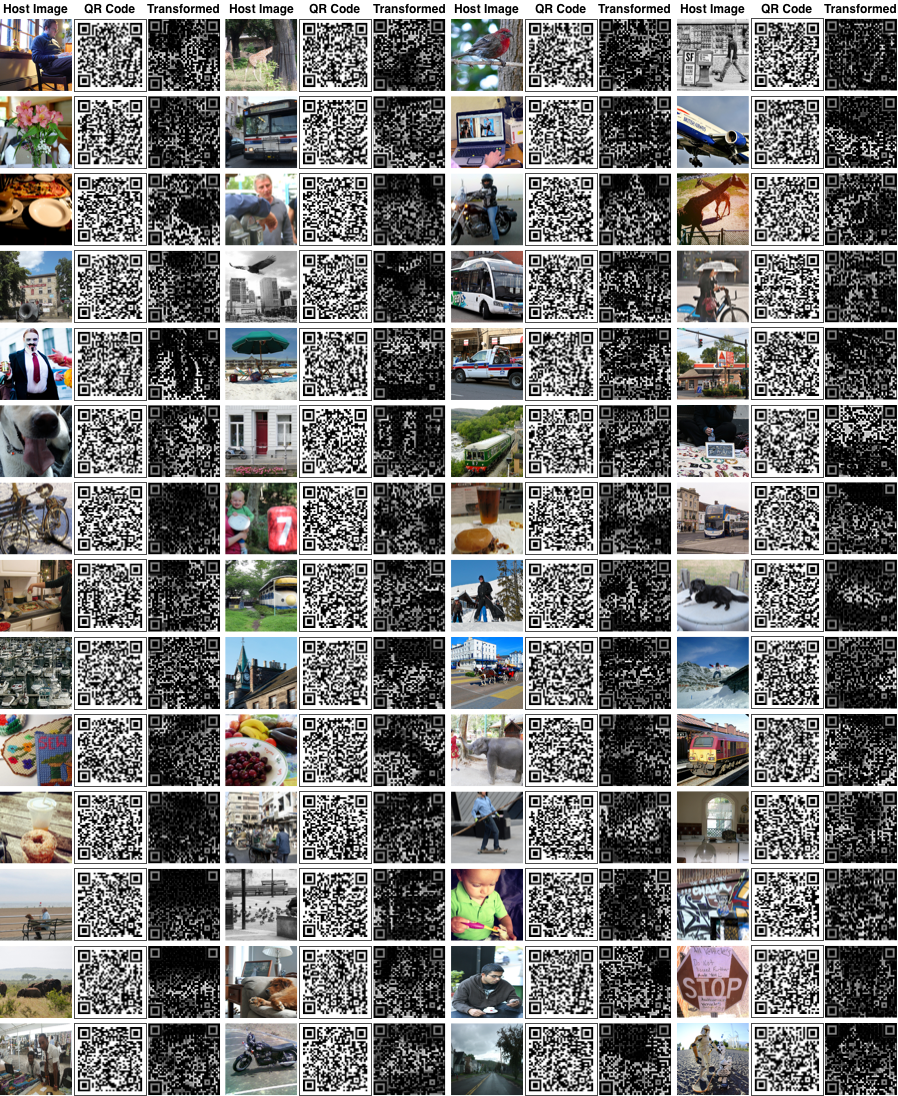}
    \caption{
        QR Code transition results, the transformed QR Codes are still identifiable.
   }
  \label{fig: sup_more_iqrt_1}
\end{figure*}

\begin{figure*}[]
    \centering
    \includegraphics[width=1.0\linewidth]{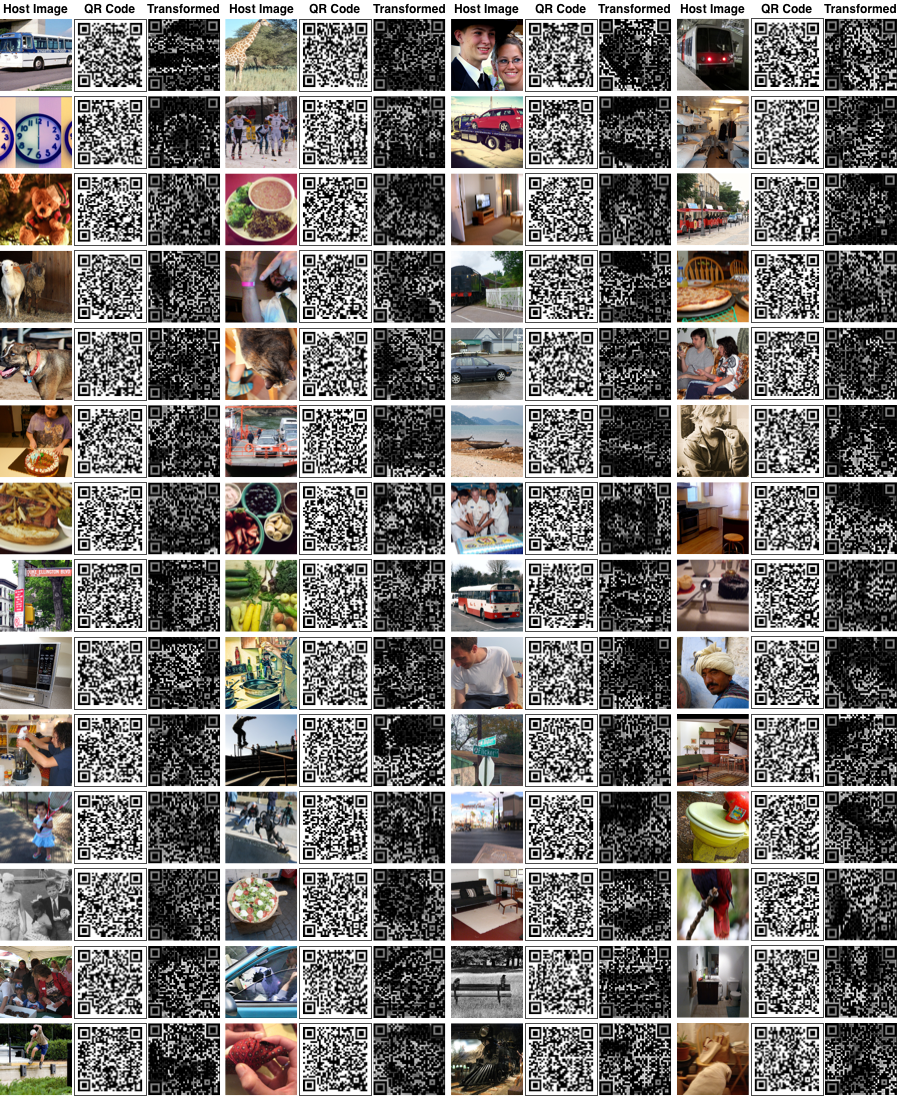}
    \caption{
        QR Code transition results, the transformed QR Codes are still identifiable.
   }
  \label{fig: sup_more_iqrt_2}
\end{figure*}

\textbf{IQRT Results} More QR Code transition results~(corresponding to \cref{fig: qr_trans}) are provided in \cref{fig: sup_more_iqrt_1} - \cref{fig: sup_more_iqrt_2}.

\begin{figure*}[]
    \centering
    \includegraphics[width=1.0\linewidth]{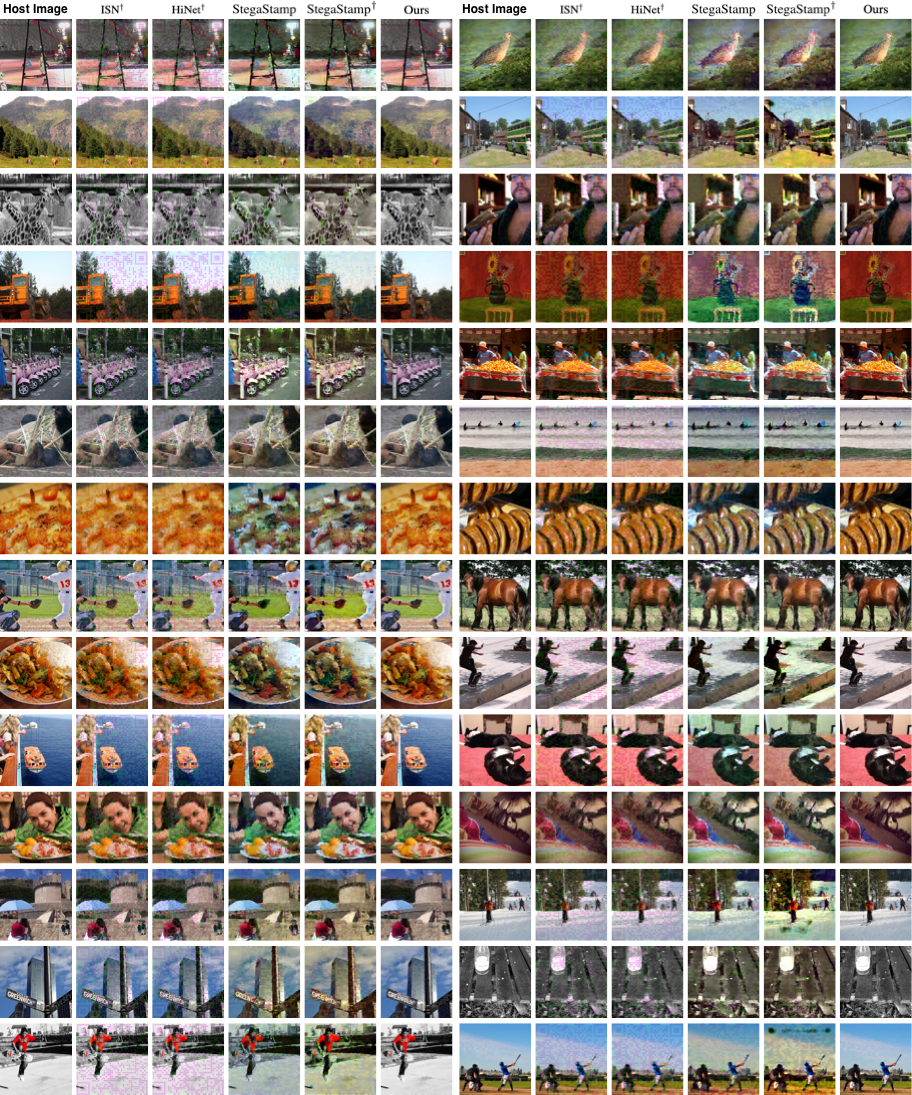}
    \caption{
        Stego images generated by different methods. Zoom in for better observation.
   }
  \label{fig: sup_more_stego_1}
\end{figure*}

\begin{figure*}[]
    \centering
    \includegraphics[width=1.0\linewidth]{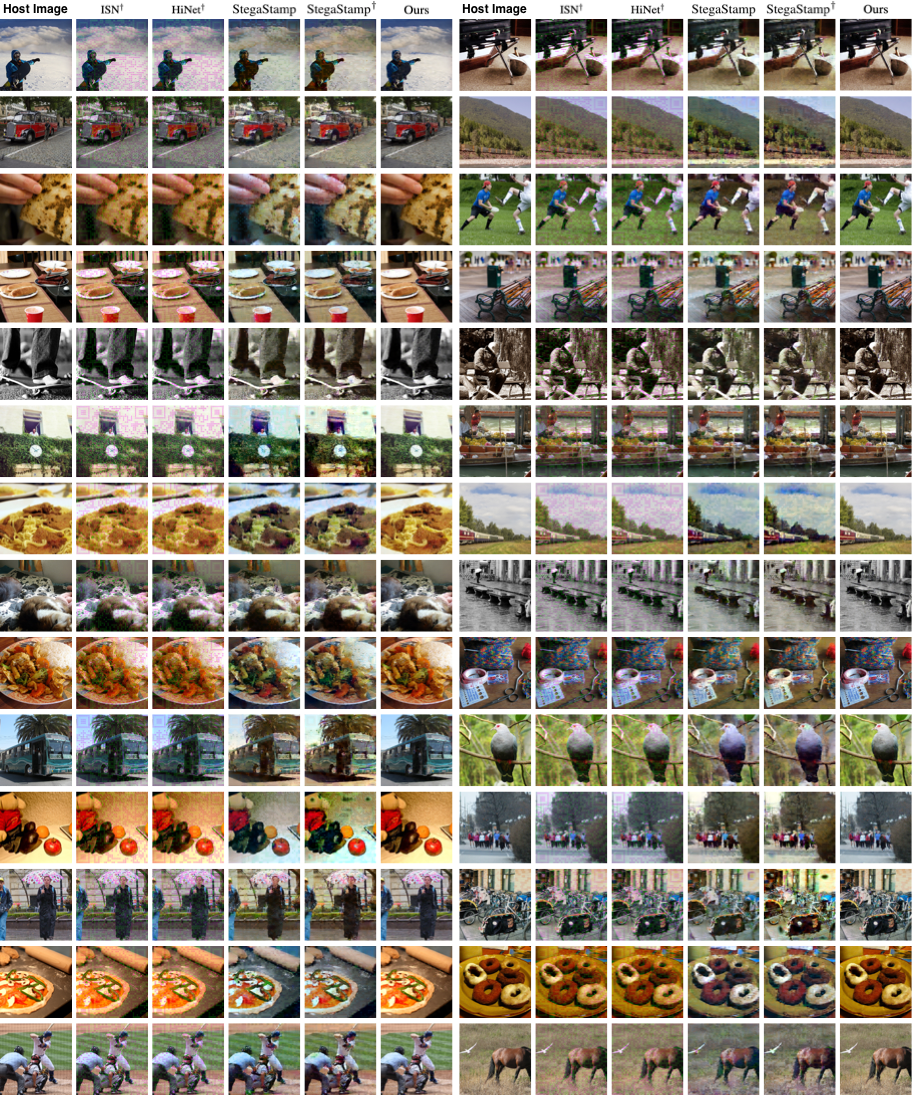}
    \caption{
        Stego images generated by different methods. Zoom in for better observation.
   }
  \label{fig: sup_more_stego_2}
\end{figure*}

\begin{figure*}[]
    \centering
    \includegraphics[width=1.0\linewidth]{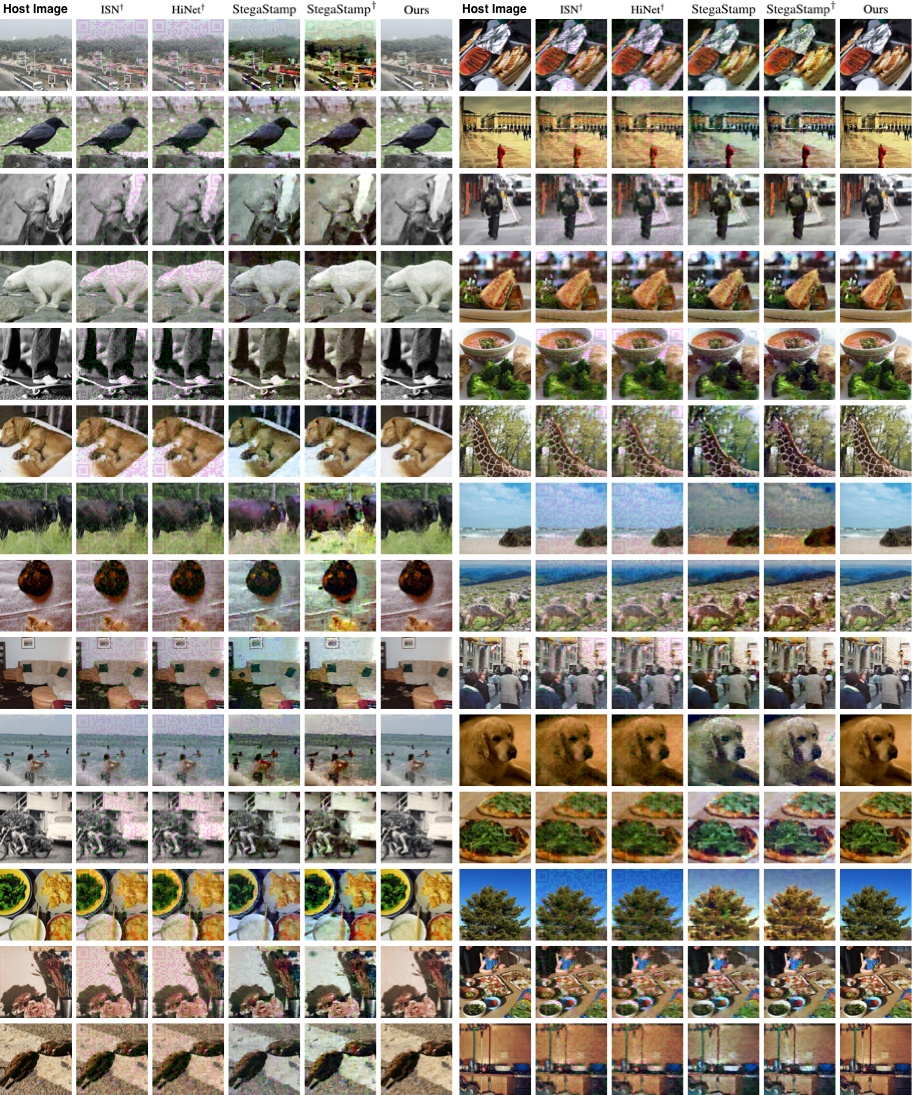}
    \caption{
        Stego images generated by different methods. Zoom in for better observation.
   }
  \label{fig: sup_more_stego_3}
\end{figure*}

\textbf{Steganography Results} More stego images generated by different methods~(corresponding to \cref{fig: evl_quality_more}) are provided in \cref{fig: sup_more_stego_1} - \cref{fig: sup_more_stego_3}.

\begin{figure*}[]
    \centering
    \includegraphics[width=1.0\linewidth]{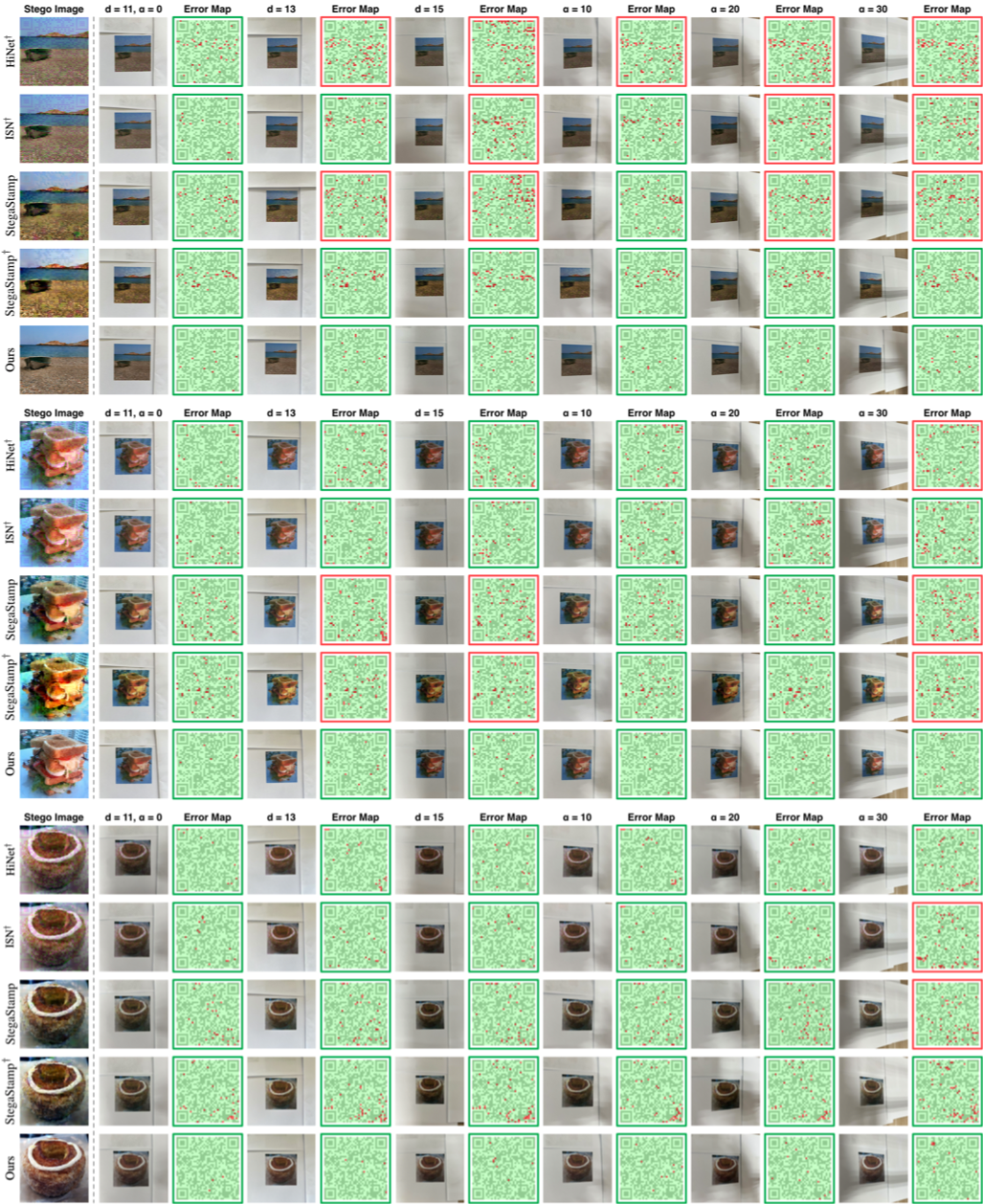}
    \caption{
        Decoding results under different shooting distances and angles. QR Codes with green borders can be successfully identified while those with red borders cannot. Zoom in for better observation.
   }
  \label{fig: sup_more_shooting_group_1}
\end{figure*}

\begin{figure*}[]
    \centering
    \includegraphics[width=1.0\linewidth]{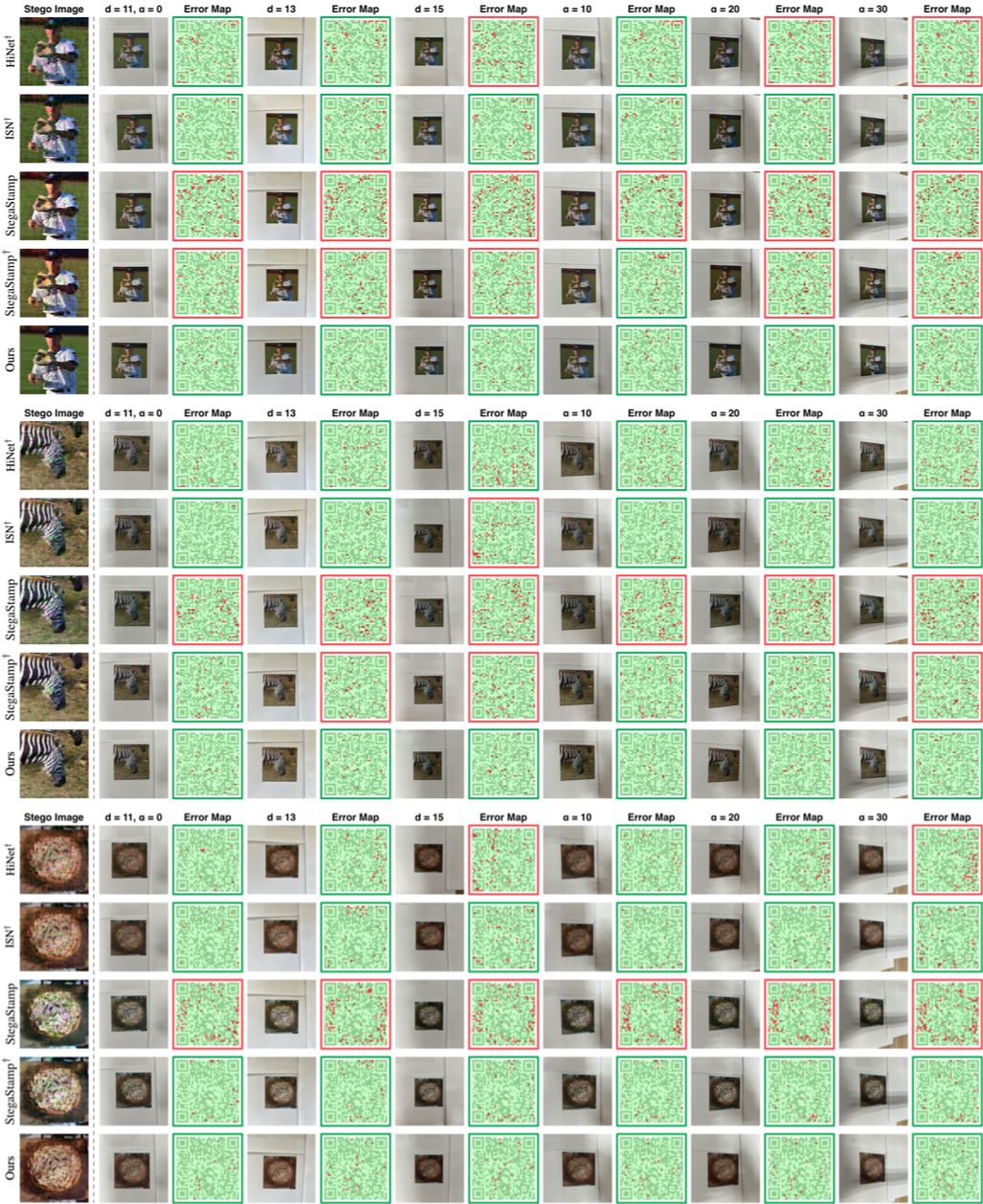}
    \caption{
        Decoding results under different shooting distances and angles. QR Codes with green borders can be successfully identified while those with red borders cannot. Zoom in for better observation.
   }
  \label{fig: sup_more_shooting_group_2}
\end{figure*}

\begin{figure*}[]
    \centering
    \includegraphics[width=1.0\linewidth]{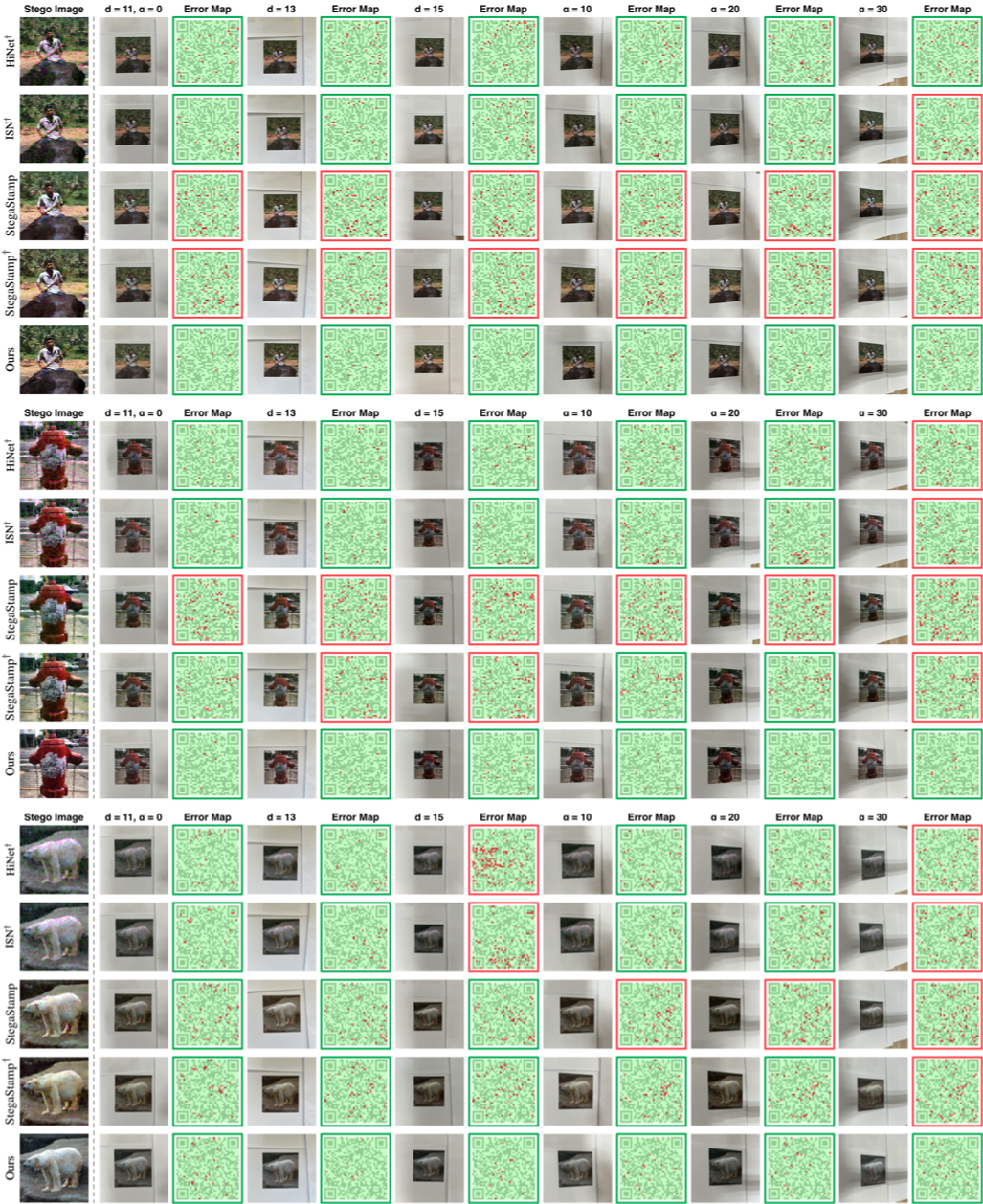}
    \caption{
        Decoding results under different shooting distances and angles. QR Codes with green borders can be successfully identified while those with red borders cannot. Zoom in for better observation.
   }
  \label{fig: sup_more_shooting_group_3}
\end{figure*}

\textbf{Print-Proof Robustness} More decoding results under different shooting situations~(corresponding to \cref{fig: evl_shooting}) are provided in \cref{fig: sup_more_shooting_group_1} - \cref{fig: sup_more_shooting_group_3}.

\begin{figure*}[]
    \centering
    \includegraphics[width=0.97\linewidth]{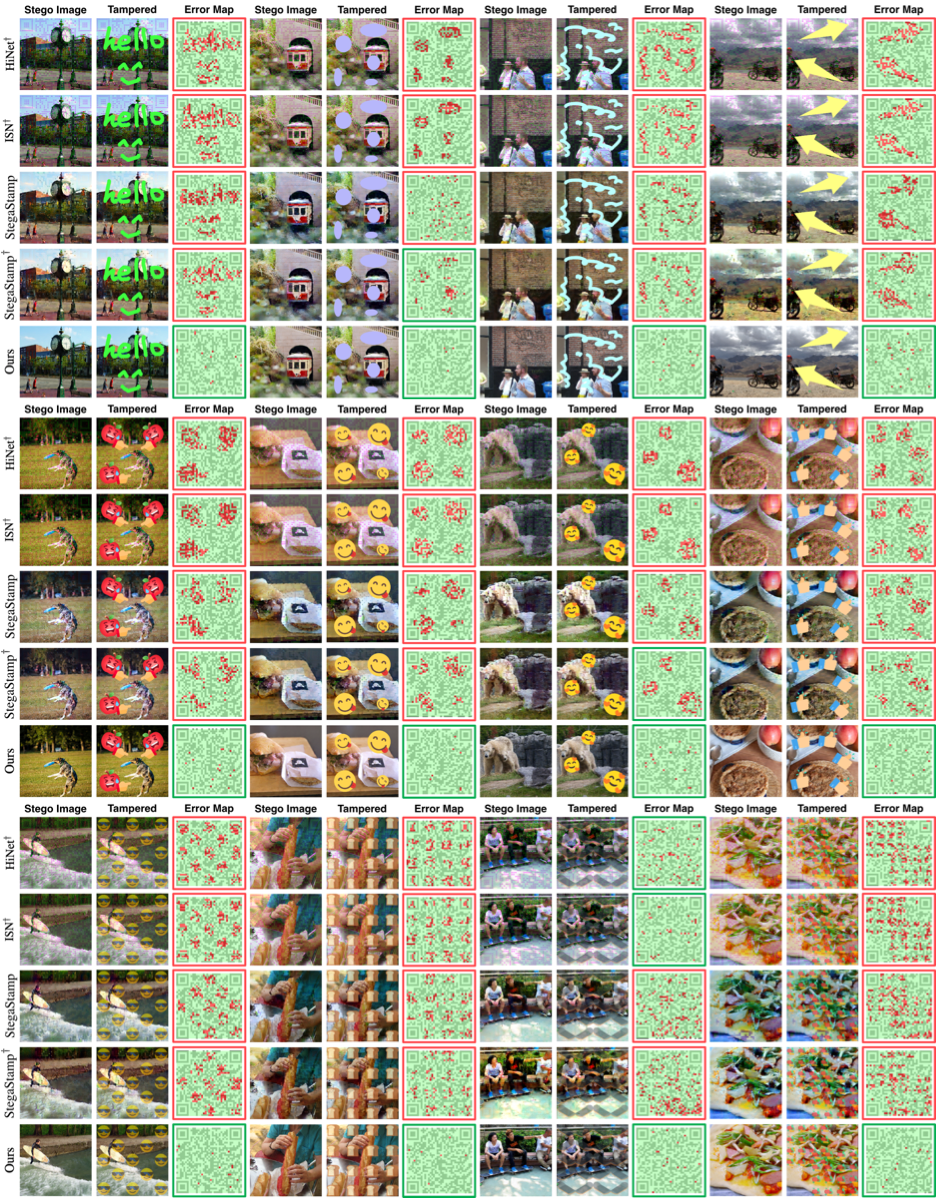}
    \caption{
        Decoding results under image tampering. QR Codes with green borders can be successfully identified while those with red borders cannot. Zoom in for better observation.
   }
  \label{fig: sup_more_tamper_group_1}\vspace{-12pt}
\end{figure*}

\textbf{Anti-Tampering} More decoding results under image tampering~(corresponding to \cref{fig: sup_tamper}) are provided in \cref{fig: sup_more_tamper_group_1}.

\begin{figure*}[]
    \centering
    \includegraphics[width=0.97\linewidth]{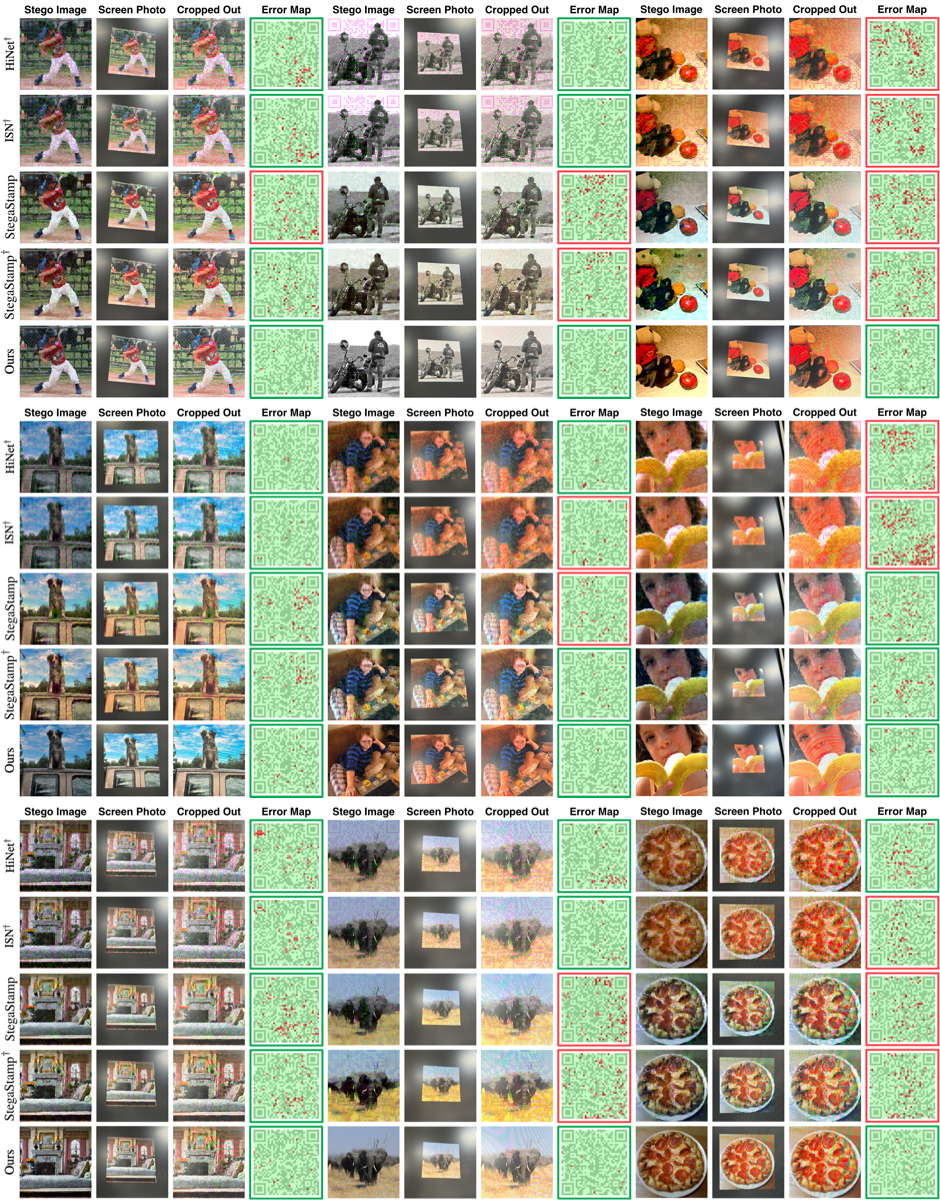}
    \caption{
        Stego images and decoding results under the distortion of light field messaging. QR Codes with green borders can be successfully identified while those with red borders cannot. Zoom in for better observation.
   }
  \label{fig: sup_more_lfm_group_1}\vspace{-8pt}
\end{figure*}

\textbf{Anti-Light Field Messaging} More decoding results under light field messaging~(corresponding to \cref{fig: sup_lfm}) are provided in \cref{fig: sup_more_lfm_group_1}.

